%% file: main.tex
\documentclass[10pt,twocolumn,letterpaper]{article}

\usepackage[pagenumbers]{cvpr} 

\input{preamble}

\definecolor{cvprblue}{rgb}{0.21,0.49,0.74}
\usepackage[pagebackref,breaklinks,colorlinks,allcolors=cvprblue]{hyperref}
\usepackage{multirow}
\usepackage[table]{xcolor}
\usepackage{array}
\usepackage{booktabs}
\usepackage{graphicx}
\usepackage{xcolor}
\usepackage{amssymb}
\usepackage[most]{tcolorbox}

\usepackage{tikz}
\usetikzlibrary{positioning,arrows.meta,calc}

\newcommand{\yes}{\textcolor{green!55!black}{\checkmark}}
\newcommand{\no}{\textcolor{red!80!black}{$\times$}}
\newcommand{\na}{--}

\definecolor{NearColor}{HTML}{EFF7F4}
\definecolor{MidColor}{HTML}{FCF5EE}
\definecolor{FarColor}{HTML}{F3F4FA}

\def\confName{CVPR}
\def\confYear{2027}

\title{ActListen: Acoustic World Models for 3D LiDAR Sound-Source Segmentation}

\author{
Haozhuang Chi$^{*}$ \quad
Soumyaratna Debnath$^{*}$ \quad
Haoyu Li\\
Taorui Wang \quad
Chen Lv$^{\dagger}$ \quad
Addison Lin Wang$^{\dagger}$\\
Nanyang Technological University\\
{\small\ttfamily lyuchen@ntu.edu.sg \quad linwang@ntu.edu.sg}\\
{\small $^{*}$Equal contribution. \quad
$^{\dagger}$Corresponding authors.}
}

\begin{document}

\twocolumn[{
\renewcommand\twocolumn[1][]{#1}%

\begin{tikzpicture}[remember picture, overlay]

\node[anchor=north west, inner sep=0pt] (empact)
at ($(current page.north west)+(0.72in,-0.43in)$)
{
    \includegraphics[
        width=3.5cm,
        trim=3 3 3 3,
        clip
    ]{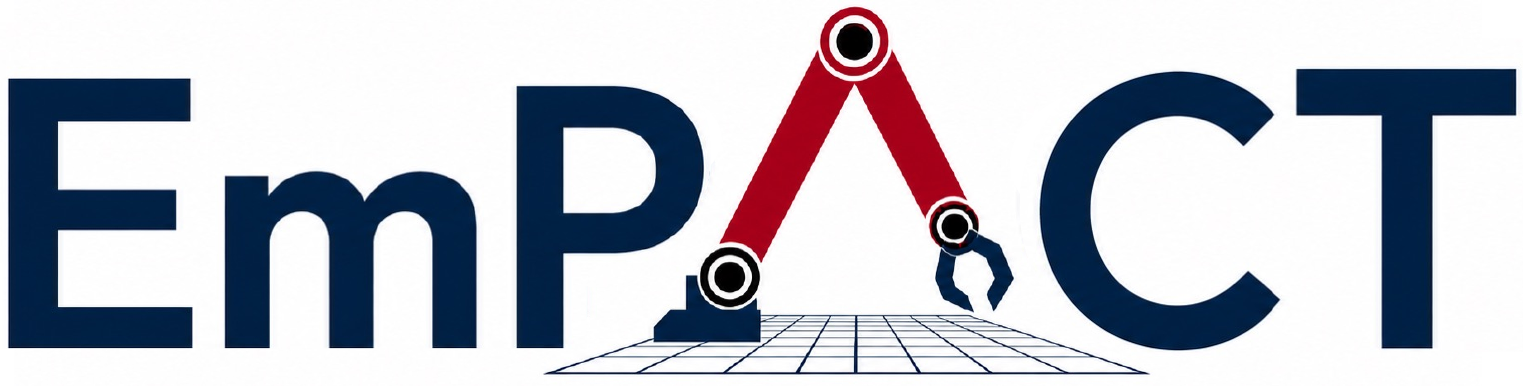}
};

\node[anchor=north east, inner sep=0pt]
at ($(current page.north east)+(-0.72in,-0.43in)$)
{
    \includegraphics[
        width=4.5cm,
        trim=3 3 3 3,
        clip
    ]{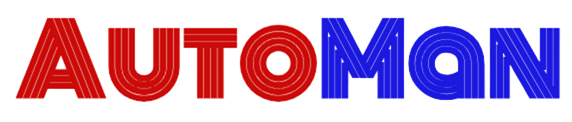}
};

\draw[
    black!20,
    line width=0.35pt
]
($(current page.north west)+(0.72in,-0.98in)$)
--
($(current page.north east)+(-0.72in,-0.98in)$);
\end{tikzpicture}

\vspace{-20pt}
\maketitle

\captionsetup{font=small,hypcap=false}

\begin{center}
    \vspace{-10pt}
    \includegraphics[width=0.95\textwidth]
    {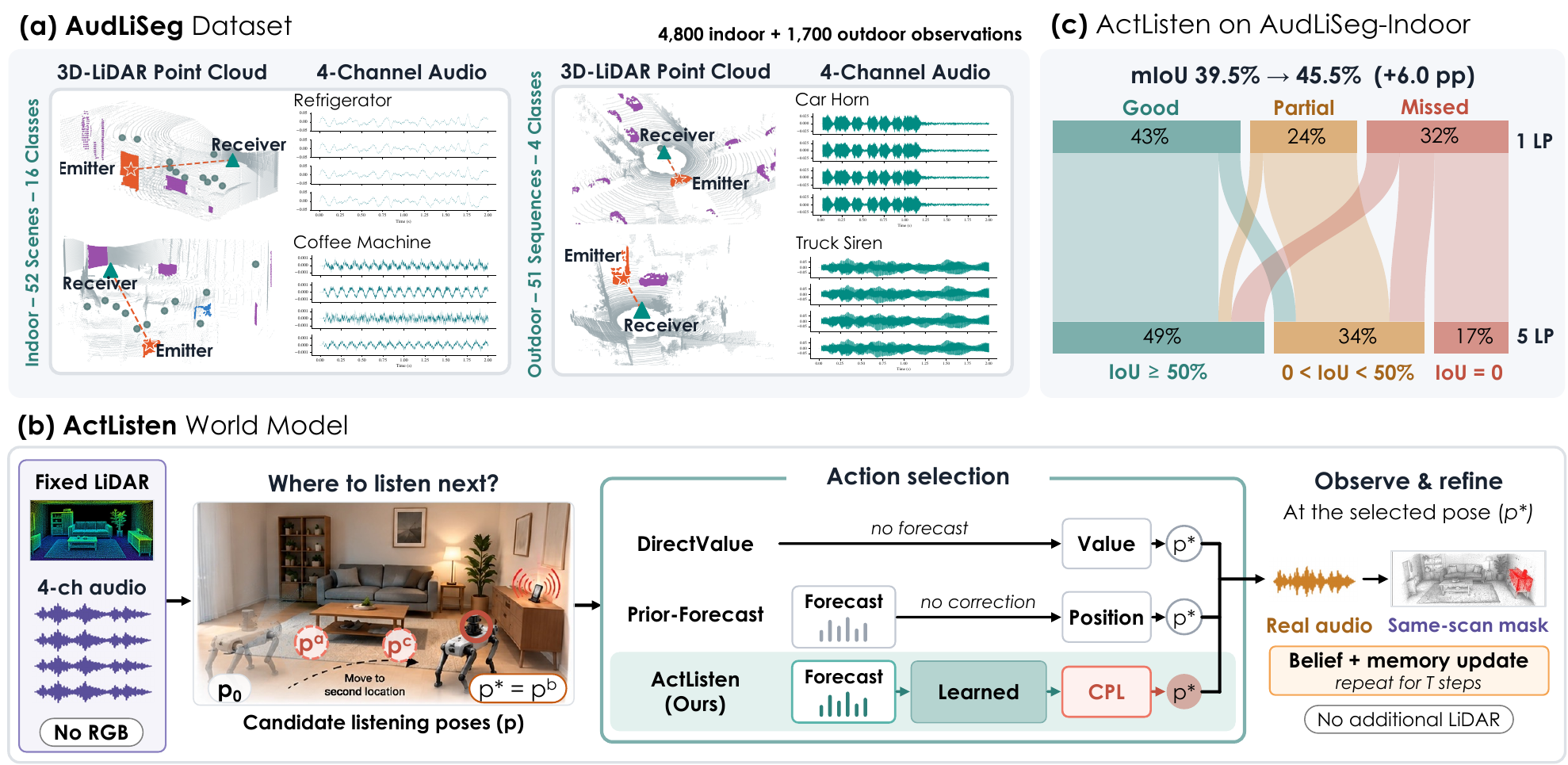}

    \vspace{-5pt}

    \captionof{figure}{
    \textbf{Overview of AudLiSeg and ActListen.}
    \textbf{(a)} AudLiSeg comprises indoor and outdoor 
    LiDAR scenes paired with four-channel audio recordings 
    of sound-emitting objects.
    \textbf{(b)} ActListen uses a geometry-conditioned 
    acoustic world model and Closed-Loop Predictive 
    Listening (CPL) to select informative listening 
    positions and iteratively refine source segmentation.
    \textbf{(c)} Active listening improves segmentation 
    on AudLiSeg-Indoor, increasing mIoU from 39.5\% 
    to 45.46\% over five listening positions while 
    reducing missed detections.}

    \label{fig:teaser}
\end{center}
\vspace{5pt}
}]

\input{sections/0_abstract}   

\input{sections/1_introduction}

\input{sections/2_related_works}
\input{sections/3.1_AudLiSeg}
\input{sections/3.2_ActListen}
\input{sections/4_experiments}
\input{sections/5_conclusion}

{
    \small
    \bibliographystyle{ieeenat_fullname}
    \bibliography{main}
}

\clearpage
\appendix
\input{suppl-main}















\end{document}

%% file: preamble.tex
\usepackage{bibunits}
\defaultbibliographystyle{ieeenat_fullname}
\defaultbibliography{main}

\makeatletter

\def\@startbibunit{%
  \global\let\@startbibunitorrelax\relax
  \global\let\@finishbibunit\@finishstartedbibunit
  \global\advance\@bibunitauxcnt 1
  \if@filesw
    {\endlinechar-1
     \makeatletter
     \@input{\@bibunitname.aux}}%
    \immediate\openout\@bibunitaux\@bibunitname.aux
    \immediate\write\@bibunitaux{\string\bibstyle{\@localbibstyle}}%
  \fi
}

\def\@finishstartedbibunit{%
  \if@filesw
    \immediate\closeout\@bibunitaux
  \fi
  \ifx\bibunits@rerun@warning\@empty
    \if@filesw
      {\endlinechar-1
       \makeatletter
       \let\@namedef\@xtestdef
       \@input{\@bibunitname.aux}}%
    \fi
  \fi
}

\makeatother

%% file: sections/0_abstract.tex
\begin{abstract}

Sound source grounding connects acoustic observations to the physical objects producing them. 
LiDAR provides precise 3D geometry, wide spatial coverage, and illumination-independent sensing, yet its potential for sound source grounding remains largely unexplored. 
To address this gap, we introduce \textbf{AudLiSeg}, the first benchmark for audio-prompted LiDAR segmentation, comprising 4,800 indoor observations across 52 HM3D scenes and 1,700 outdoor observations on KITTI/SemanticKITTI, with four-channel spatial audio, multi-pose sensing, and point-level active-source annotations.
Consequently, we propose \textbf{ActListen}, an active framework for identifying and segmenting sounding objects without RGB input.
At its core, ActListen employs \textbf{Acoustic Hypothesis Grounding} module to construct candidate object masks, emission hypotheses, and initial acoustic states.
Building on these hypotheses, \textbf{Closed-Loop Predictive Listening} leverages a geometry-conditioned acoustic world model that combines physical sound propagation with learned representations to predict acoustic observations, select informative listening positions, and update source beliefs from acquired recordings.
The final segmentation is produced on the original LiDAR point cloud without additional scans.
We evaluate ActListen on AudLiSeg against adapted sound source localization and multimodal grounding baselines.
Using five listening positions, ActListen achieves an average mIoU of \textbf{45.46\%} on the unseen AudLiSeg-Indoor test set, corresponding to a \textbf{9.75\% relative improvement} over random pose selection.
On the Outdoor set, ActListen achieves \textbf{31.76\%} mIoU. Project Page : \href{https://empactlab.github.io/ActListen/}{https://empactlab.github.io/ActListen/}.
\vspace{-50pt}
\end{abstract}

%% file: sections/1_introduction.tex
\input{tables/dataset_comparision}

\vspace{-10pt}
\section{Introduction}
\label{sec:intro}
Sound source grounding is fundamental to multimodal scene understanding, complementing semantic recognition~\cite{chi2025vlm} by associating acoustic observations with the physical objects producing them.
Existing audio-visual segmentation methods establish this correspondence in 2D images~\cite{malard2026taco, guo2025audio}, while recent 3D approaches extend it to sound source localization and scene grounding~\cite{sokolov20243d,he2025soundloc3d}.
However, recognizing a sound's category or estimating its location does not necessarily identify the specific object instance responsible for it~\cite{xuan2024robust,um2025object}.
LiDAR offers precise 3D geometry, wide spatial coverage, and illumination-independent sensing~\cite{behley2019semantickitti,caesar2020nuscenes}, making it a promising modality for grounding sounds to individual objects in 3D.
Despite these advantages, existing research has primarily focused on image-based audio-visual segmentation or 3D sound localization, leaving audio-prompted LiDAR instance segmentation largely unexplored and without a dedicated benchmark.

To address this gap, we introduce \textbf{AudLiSeg}, the \textbf{\textit{first}} benchmark for audio-prompted LiDAR segmentation that enables systematic evaluation of sound-producing object grounding in 3D point clouds (Fig.~\ref{fig:teaser}\textcolor{cvprblue}{a}). 
AudLiSeg combines LiDAR geometry, instance-level annotations, and spatial multichannel audio across simulated indoor and real outdoor LiDAR scenes with synthesized acoustics. 
Beyond single-view segmentation, the benchmark supports multiple listening positions, enabling evaluation of both passive grounding and active acoustic evidence acquisition.

However, identifying a sound-producing object from a single observation remains challenging~\cite{chakraborty2014source,jalayer2025review}. Similar acoustic signatures, reflections, and complex sound propagation can make multiple source instances consistent with the same recording~\cite{jalayer2025review,braun2019acoustic}. Even with precise LiDAR geometry, such ambiguities cannot always be resolved through feature fusion alone. Acquiring audio from different positions can provide complementary spatial evidence~\cite{schult2015information,yamada2023placement}, motivating a key question: \emph{where should an agent listen next to accurately identify the sound-producing object?}

To address this challenge, we propose \textbf{ActListen}, a \textbf{novel} active audio-prompted LiDAR segmentation framework that identifies and segments sound-producing objects without RGB input (Fig.~\ref{fig:teaser}\textcolor{cvprblue}{b}). Unlike existing methods that focus on navigation~\cite{chen2021semantic} or sound separation~\cite{majumder2021move2hear}, ActListen actively acquires acoustic evidence to improve object-level segmentation. At its core, ActListen employs \textbf{Acoustic Hypothesis Grounding} module to construct candidate object masks, emission hypotheses, and initial acoustic states.
Building on these hypotheses, \textbf{Closed-Loop Predictive Listening} leverages a geometry-conditioned acoustic world model that combines physical sound propagation with learned representations to predict acoustic observations, select informative listening positions, and update source beliefs from acquired recordings.
The final segmentation is produced on the original LiDAR point cloud.

\begin{figure*}[t!]
  \centering
  \includegraphics[width=1.0\linewidth]{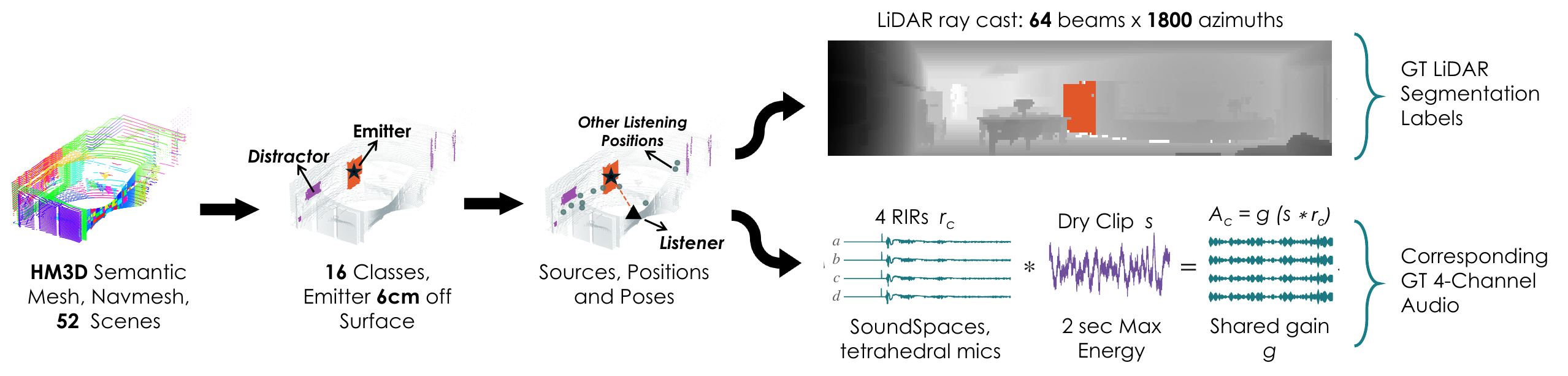}
\vspace{-23pt}
\caption{\textbf{AudLiSeg-Indoor benchmark construction.} Semantically annotated HM3D scenes are used to generate 360$^\circ$ LiDAR scans with ground-truth instance masks. SoundSpaces simulates acoustic propagation using a tetrahedral microphone array.}
\vspace{-15pt}
  \label{fig:dataset_pipeline}
\end{figure*}




    
\noindent In summary, our main contributions are:
\begin{itemize}
    \item To the best of our knowledge, we are the \emph{first} to address active audio-prompted LiDAR segmentation using only LiDAR and acoustic observations, without RGB input.

\item We introduce \textbf{AudLiSeg}, a benchmark for audio-prompted LiDAR segmentation across diverse indoor and outdoor environments (\textbf{Sec.~\ref{sec:audLiSeg}, Table~\ref{tab:dataset_comparison}}).

    \item We propose \textbf{ActListen}, a world model-based active segmentation framework that predicts acoustic observations at unseen positions, selects informative listening viewpoints, and iteratively refines source segmentation without additional LiDAR scans (\textbf{Sec.~\ref{sec:actListen}}).

    \item Extensive experiments on AudLiSeg demonstrate the effectiveness of active listening against adapted localization and multimodal grounding methods.
\end{itemize}

%% file: tables/dataset_comparision.tex
\begin{table*}[t]
\caption{\textbf{Comparison with related audio-visual and
3D segmentation datasets.}
AudLiSeg combines spatial audio, LiDAR, point-level sounding-object annotations, and active listening across indoor and outdoor environments.
\yes: yes; \no: no; \na: not applicable or not reported.}
\label{tab:dataset_comparison}
\centering
\vspace{-7pt}
\resizebox{\textwidth}{!}{%
\begin{tabular}{lccccccccl}
\toprule
& \multicolumn{3}{c}{Input Modalities}
& \multicolumn{3}{c}{Evaluation}
& \multicolumn{2}{c}{Environment}
& \multirow{2}{*}{\quad\quad\quad\quad\quad\textbf{Primary Task}} \\
\cmidrule(lr){2-4}
\cmidrule(lr){5-7}
\cmidrule(lr){8-9}
\textbf{Dataset}
& \textbf{Audio} & \textbf{RGB} & \textbf{LiDAR}
& \textbf{3D Mask} & \textbf{Multi-View} & \textbf{Active}
& \textbf{Indoor} & \textbf{Outdoor}
& \\
\midrule

AVSBench~\cite{zhou2022audio}
& \yes & \yes & \no
& \no & \no & \no
& \na & \na
& 2D audio-visual segmentation \\

3DAVS-S34-O7~\cite{sokolov20243d}
& \yes & \yes & \no
& \yes & \yes & \no
& \yes & \no
& 3D audio-visual segmentation \\


SemanticKITTI~\cite{behley2019semantickitti}
& \no & \yes & \yes
& \no & \yes & \no
& \no & \yes
& LiDAR semantic segmentation \\





nuScenes~\cite{caesar2020nuscenes}
& \no & \yes & \yes
& \yes & \yes & \no
& \no & \yes
& LiDAR panoptic segmentation \\

SceneFun3D~\cite{delitzas2024scenefun3d}
& \no & \yes & \yes
& \yes & \yes & \no
& \yes & \no
& Language-guided 3D segmentation \\

\midrule
\rowcolor{blue!8}
\textbf{AudLiSeg (Ours)}
& \yes & \no & \yes
& \yes & \yes & \yes
& \yes & \yes
& \textbf{Audio-prompted LiDAR segmentation} \\

\bottomrule
\end{tabular}%
}
\end{table*}

%% file: sections/2_related_works.tex
\vspace{-2pt}
\section{Related Works}
\label{sec:related_work}
\vspace{-3pt}

\noindent\textbf{Audio-Visual and 3D Scene Understanding.}
Audio-visual scene understanding links sounds to objects and their spatial
extent. AVSBench~\cite{zhou2022audio} established pixel-level sounding-object
segmentation; TACO~\cite{malard2026taco} derives segmentation prompts through
semantically constrained factorization of frozen audio and visual representations.
AVIS~\cite{guo2025audio} identifies, segments, and tracks sounding video instances;
BiAVIS~\cite{wang2026biavis} uses binaural spatial cues to distinguish instances
of the same category. Both separate category recognition from emitting-object
identification.
In 3D, Point-Bind~\cite{guo2023pointbind} embeds point clouds, audio, images,
and language in a shared space. EchoSegnet~\cite{sokolov20243d} lifts
audio-conditioned image masks into a 3D Gaussian scene and refines them
with spatial audio. SoundLoc3D~\cite{he2025soundloc3d} combines microphone-array
measurements and RGB-D surface cues to estimate source locations and categories.
In contrast, we directly segment sound-producing instances on LiDAR point clouds without RGB input, leveraging active listening to resolve ambiguities among candidate sources.

\vspace{2pt}
\noindent\textbf{Active Perception.}
Active perception couples scene understanding with adaptive observation selection to reduce uncertainty~\cite{bajcsy1988active}.
Bayesian experimental design formalizes this through expected information gain, while Deep Adaptive Design~\cite{foster2021dad} learns sequential acquisition policies conditioned on previous observations.
In embodied environments, audio-visual navigation leverages semantic and spatial acoustic cues to locate sound sources, with memory supporting navigation when sounds become unavailable~\cite{chen2021semantic}.
MAGNet~\cite{zeng2026savnce} integrates acoustic history and self-motion for continuous navigation, while Move2Hear~\cite{majumder2021move2hear} optimizes agent motion for sound separation.
Unlike these approaches, ActListen selects informative listening positions to resolve ambiguities among candidate sound sources and refine instance segmentation on a fixed LiDAR point cloud, isolating gains from acoustic evidence without additional geometric observations.

\vspace{2pt}
\noindent\textbf{World Models for Embodied Intelligence.}
World models predict state evolution~\cite{chi2026driver,chi2026pv} and action consequences for planning and decision-making.
PlaNet~\cite{hafner2019planet} learns latent visual dynamics for planning, while WoMAP~\cite{yin2025womap} predicts dynamics and rewards for active object localization.
UniWM~\cite{dong2026towards} integrates visual prediction and planning with hierarchical memory, and LCVN~\cite{dong2026language} employs latent imagination for language-conditioned navigation.
Beyond visual prediction, acoustic world models capture how sound changes with source and receiver configurations.
Neural Acoustic Fields~\cite{luo2022naf} predict impulse responses from emitter and listener positions, while Novel-View Acoustic Synthesis~\cite{chen2023nvas} and SoundVista~\cite{chen2025soundvista} synthesize audio at unseen listening positions.
Disentangled Acoustic Fields~\cite{yin2024daf} model sound formation and propagation for source inference, while AVWM~\cite{wang2025avwm} predicts action-conditioned audiovisual observations for navigation.
Unlike these approaches, ActListen employs a geometry-conditioned acoustic world model to predict receiver-dependent acoustic evidence, guide informative listening position selection, and update source hypotheses from acquired recordings for segmentation on the original LiDAR point cloud.

%% file: sections/3.1_AudLiSeg.tex
\section{AudLiSeg Benchmark}
\label{sec:audLiSeg}
\vspace{-3pt}

\begin{figure*}[t!]
  \centering
  \includegraphics[width=1.0\linewidth]{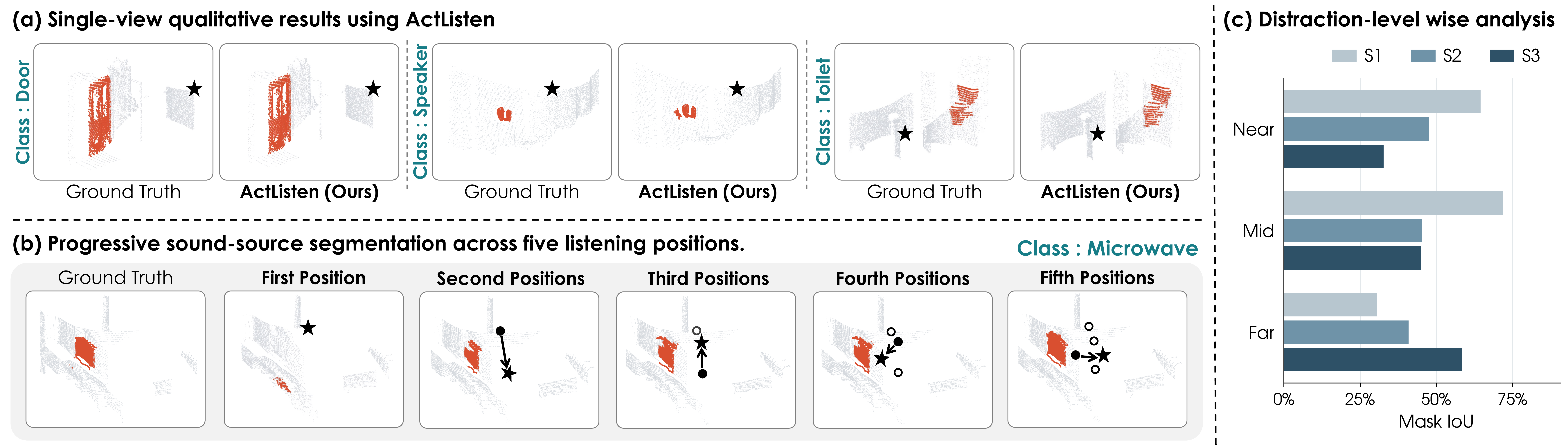}
  \vspace{-20pt}
  \caption{\textbf{Evaluation of ActListen on AudLiSeg-Indoor.}
\textbf{(a)} Single-view qualitative results across object classes.
\textbf{(b)} Progressive segmentation over five listening positions.
$\bigstar$ indicates the current listening position, $\bullet$ marks the immediately preceding position, and $\circ$ denotes all earlier positions.
\textbf{(c)} ActListen on AudLiSeg-Indoor across distractor states/levels (S1, S2, S3).}
  \label{fig:actlisten_qual}
  \vspace{-15pt}
\end{figure*}

\textbf{AudLiSeg} evaluates audio-prompted LiDAR segmentation across indoor
and outdoor scenes with simulated spatial acoustics
(Fig.~\ref{fig:teaser}\textcolor{cvprblue}{a}). Semantically annotated
Habitat-Matterport 3D (HM3D) scenes~\cite{ramakrishnan2021habitat} provide
simulated indoor LiDAR and rendered four-channel audio from 16 object
categories. SemanticKITTI/KITTI point clouds and instance
annotations~\cite{behley2019semantickitti} provide real outdoor driving geometry with recorded vehicle sounds (ESC-50) rendered by simulated propagation. Both branches use a calibrated
tetrahedral array and \emph{human-verified} point-level active-source
instance annotations. 
\textit{More details in Suppl. Mat. Sec.~\ref{supp_sec:aduliseg}}.


\vspace{2pt}
 \noindent\textbf{Indoor Branch.}
Fig.~\ref{fig:dataset_pipeline} illustrates the construction pipeline of AudLiSeg-Indoor. Simulated 64-beam, $360^\circ$ LiDAR scans are generated from HM3D semantic meshes~\cite{ramakrishnan2021habitat},
with instance labels retained for simulated returns. Each source from
\textbf{16} sound-capable categories remains fixed across \textbf{12}
navigable receiver poses. A tetrahedral four-microphone array centered at
the LiDAR origin, with $9.24\,\mathrm{cm}$ pairwise spacing, captures spatial
cues. SoundSpaces~2.0~\cite{chen2022soundspaces} simulates direct sound,
reflections, diffraction, and transmission with material-dependent acoustic
properties, yielding two seconds of synchronized four-channel audio at
$44.1\,\mathrm{kHz}$.
The indoor subset has \textbf{4,800 observations} of \textbf{400 source
instances} in \textbf{52 scenes}, split into 3,264/768/768 training/validation/test samples with disjoint scenes and disjoint source recordings across splits. Source--receiver
distance tiers are \textit{Near-Field} ($d\leq3\,\mathrm{m}$),
\textit{Mid-Field} ($3<d\leq5\,\mathrm{m}$), and \textit{Far-Field}
($d>5\,\mathrm{m}$), with 2,735/1,416/649 observations, respectively.
Visible sound-capable distractors define ambiguity tiers: none ($S1$), only
different-class objects ($S2$), or at least one same-class object ($S3$).
These contain 94/3,681/1,025 observations, respectively, each with one
active source.

\vspace{2pt}
\noindent\textbf{Outdoor Branch.}
KITTI HDL-64E point clouds, SemanticKITTI instance
annotations~\cite{behley2019semantickitti}, and sensor poses provide outdoor
geometry. Three sound recordings (engine, horn, and siren) from ESC-50~\cite{piczak2015esc}
are assigned at labeled vehicle locations across four vehicle classes from SemanticKITTI. An analytical model of direct
propagation and planar ground reflection generates impulse responses at
\textbf{12} receiver poses. Convolution with a two-second source segment at
each microphone of the same tetrahedral array yields synchronized
four-channel audio at $44.1\,\mathrm{kHz}$. The \textbf{1,700 observations}
span \textbf{370 source groups} and \textbf{51 sequence windows}, with
sequence-disjoint training/validation/test splits of 1,088/326/286.
\textit{Near-Field} ($d\leq5\,\mathrm{m}$), \textit{Mid-Field}
($5<d\leq9\,\mathrm{m}$), and \textit{Far-Field} ($d>9\,\mathrm{m}$)
contain 542/593/565 observations, respectively.

%% file: sections/3.2_ActListen.tex
\section{ActListen}
\label{sec:actListen}
\vspace{-2pt}

\begin{figure*}[t!]
  \centering
  \includegraphics[width=1.0\linewidth]{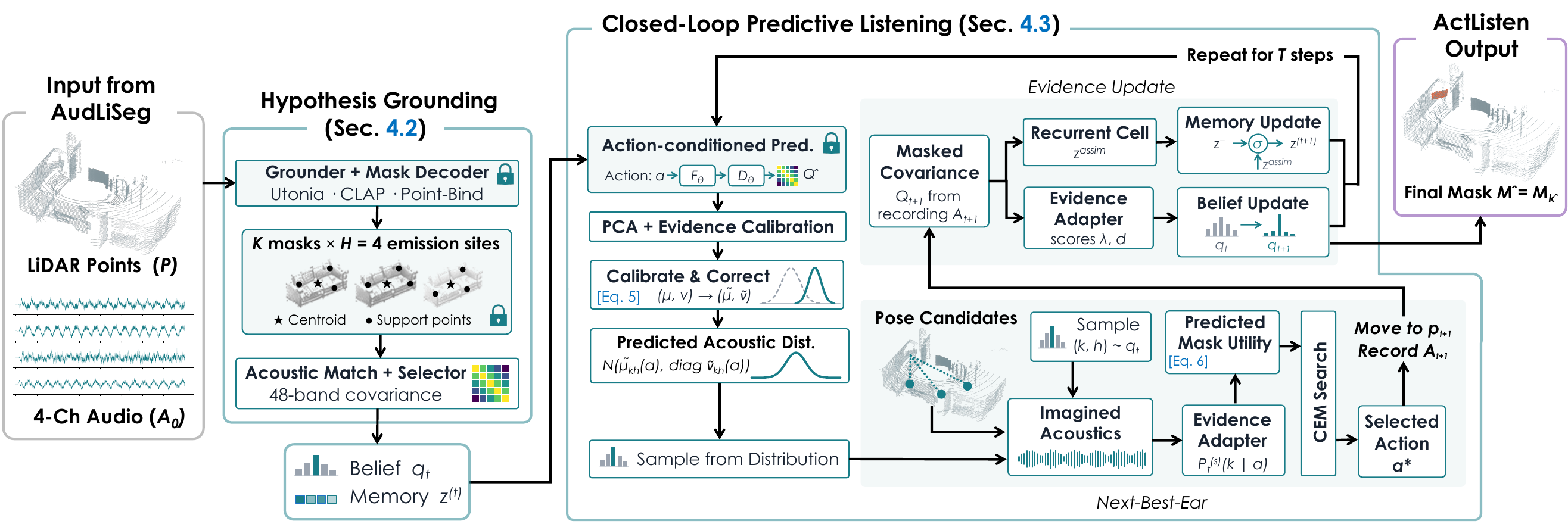}
\vspace{-15pt}
\caption{\textbf{Overview of ActListen's Pipeline.} An acoustic world model for audio-prompted LiDAR segmentation without RGB.
\textit{Additional methodological details are provided in in Suppl. Mat. Sec.~\ref{supp_sec:actlisten}.}
}
  \label{fig:actlisten_method}
\end{figure*}

Given a LiDAR scene and multichannel audio, ActListen identifies and segments the sound-producing object without relying on RGB, using a closed-loop process of acoustic prediction, active pose selection, and evidence refinement (Fig.~\ref{fig:actlisten_method}). Starting from initial object and emission-site hypotheses, the acoustic world model predicts observations at candidate receiver poses to progressively refine segmentation using acquired acoustic evidence.

\subsection{Problem Formulation}
\label{sec:problem}
\vspace{-2pt}

Given reference LiDAR points $\mathcal{P}=\{\mathbf{x}_n\}_{n=1}^{N}$ and a
four-channel recording $\mathbf{A}_0$ at pose $\mathbf{p}_0$, ActListen segments
the sounding object on these points. The scene and source are static;
receiver poses and microphone calibration are available, and a navigability
map constrains motion. No RGB or further LiDAR is acquired. Action
$a_t=(\mathbf{d}_t,\mathbf{R}_t)$ specifies the next receiver transform relative
to its current pose; execution yields $\mathbf{A}_{t+1}$.

A fixed bank contains $K$ candidate masks $\{M_k\}$, each with $H$ emission
locations. Their joint belief $q_t(k,h)$ gives candidate marginal
$p_t(k)=\sum_h q_t(k,h)$ and conditional emission-site belief
$\rho_t(h\mid k)=q_t(k,h)/p_t(k)$. After $T$ additional recordings (total
$T+1$ observations) we predict
{\setlength\abovedisplayskip{2pt}\setlength\belowdisplayskip{2pt}
\begin{equation}
    \hat M=M_{\hat k},\qquad \hat k=\arg\max_k p_T(k).
    \label{eq:mask_output}
\end{equation}}
The candidate masks remain fixed throughout sensing.

\subsection{Acoustic Hypothesis Grounding}
\label{sec:hypothesis_grounding}

\noindent\textbf{Object and Emission Hypotheses.}
We encode the reference LiDAR geometry using Utonia~\cite{zhang2026utonia}. 
Audio-conditioned point scores, spatial acoustic cues, semantic compatibility, and geometric confidence are combined to select spatially distributed prompts for the mask decoder. 
Multiple mask thresholds generate candidate object extents, after which small and duplicate proposals are removed. 
The resulting masks may overlap or represent alternative extents of the same object. 
Each candidate is associated with $H=4$ geometric emission hypotheses: a point-count-weighted centroid and three spatially distributed support points, which serve as candidate source locations rather than measured emission positions. 
Point-Bind object features and CLAP audio features provide additional semantic compatibility cues~\cite{guo2023pointbind,wu2022clap}.

\vspace{2pt}
\noindent\textbf{Initial acoustic evidence.}
Each recording is represented by band-averaged, trace-normalized spatial
covariances $\mathbf Q^{\mathrm{obs}}_{tb}\in\mathbb C^{4\times4}$,
with $B=48$ bands and an active-band mask $m_{tb}$.
For each emission hypothesis, the acoustic model pools global geometry,
local source context, and points near the source--receiver segment.
It initializes a latent state and predicts a covariance
$\mathbf Q_{khb}$ and scalar log error scale $\ell_{khb}$.
The covariance is a learned convex mixture of an analytical direct-path
prediction and a positive-semidefinite, trace-normalized learned component.
The initial acoustic discrepancy is
{\setlength\abovedisplayskip{2pt}\setlength\belowdisplayskip{2pt}
\begin{equation}
    E_{khb}=e^{-\ell_{khb}}
    \|\mathbf Q_{khb}-\mathbf Q^{\mathrm{obs}}_{0b}\|_F^2+\ell_{khb}.
    \label{eq:actlisetn_acoustic}
\end{equation}}
A selector combines the weighted discrepancies, semantic evidence, and
proposal scores to produce $b_k$ and $\rho_0(h\mid k)$. The initial belief is
{\setlength\abovedisplayskip{2pt}\setlength\belowdisplayskip{2pt}
\begin{equation}
    q_0(k,h)=\operatorname{softmax}_k(\alpha\mathbf b)\,\rho_0(h\mid k),
    \label{eq:initial-belief}
\end{equation}}
Here, $\alpha>0$ is a learned prior scale. A recurrent cell encodes the
initial covariance residual into $\mathbf z_{kh}^{(0)}\in\mathbb R^{128}$.
Together with $q_0$, these memories initialize the listening loop.
The reference grounder and base acoustic model are frozen in subsequent
training stages.

\subsection{Closed-Loop Predictive Listening}
\label{sec:predictive_listening}
\vspace{-2pt}
Starting from the current object-site belief $q_t$ and acoustic memories $\{\mathbf z_{kh}^{(t)}\}$, ActListen first uses \textit{action-conditioned prediction} to forecast the acoustic observations expected at candidate receiver poses. \textit{Next-Best-Ear (NBE)} then evaluates how these predicted observations would affect the source-mask decision and selects the most informative pose. The acquired recording updates both the belief and acoustic memory before the next listening step.

\noindent\textbf{Action-Conditioned Prediction.}
Given a candidate receiver action $a$, the acoustic world model predicts how each source hypothesis would be observed from the corresponding new pose. It transforms the source coordinates and propagates the associated acoustic memory as
{\setlength\abovedisplayskip{2pt}\setlength\belowdisplayskip{2pt}
\begin{equation}
\begin{aligned}
 (\mathbf z^-_{kh}(a),\mathbf s'_{kh}(a))
    &=F_\theta(\mathbf z_{kh}^{(t)},\mathbf s_{kh}^{(t)},a),\\
 (\mathbf Q_{kh}(a),\boldsymbol\ell_{kh}(a))
    &=D_\theta(\mathbf z^-_{kh}(a),\mathbf s'_{kh}(a)).
\end{aligned}
\label{eq:acoustic_transition}
\end{equation}}
Here, $F_\theta$ applies the known receiver motion and advances the latent acoustic state, while $D_\theta$ predicts the corresponding multiband covariance $\mathbf Q_{kh}(a)$ and uncertainty $\boldsymbol\ell_{kh}(a)$. These predictions are hypothetical and do not modify the retained memory until the selected pose is actually visited.

To obtain a compact distribution for acoustic imagination, we project the predicted covariances into a 128-dimensional PCA space and calibrate their mean and diagonal variance,
$\boldsymbol\mu_{kh}(a)$ and $\boldsymbol v_{kh}(a)$, using an evidence adapter. A lightweight observation model then refines these predictions using the current acoustic memory, latest residual, source hypothesis, candidate action, and belief:
{\setlength\abovedisplayskip{2pt}\setlength\belowdisplayskip{2pt}
\begin{equation}
\begin{aligned}
 \tilde{\boldsymbol\mu}_{kh}(a)
    &=\boldsymbol\mu_{kh}(a)+0.5\tanh\boldsymbol\Delta_{kh}(a),\\
 \tilde{\boldsymbol v}_{kh}(a)
    &=\boldsymbol v_{kh}(a)\odot
      \exp\!\left(0.7\tanh\boldsymbol\Gamma_{kh}(a)\right).
\end{aligned}
\label{eq:predictive_acoustics}
\end{equation}}
The resulting distribution represents plausible acoustic observations at the candidate pose and is used by NBE to evaluate where to listen next. The evidence adapter separately scores both imagined and measured observations; for measured recordings, inactive frequency bands are masked before evaluating the acoustic evidence.

\vspace{2pt}
\noindent\textbf{Next-Best-Ear (NBE) Selection.}
NBE selects listening poses based on their expected effect on the final mask decision, rather than acoustic variation or geometric coverage alone. For each feasible action $a$, it samples a source hypothesis $(k_s,h_s)\sim q_t$ and draws an imagined acoustic feature from
$\mathcal N(\tilde{\boldsymbol\mu}_{k_s h_s}(a),
\operatorname{diag}\tilde{\boldsymbol v}_{k_s h_s}(a))$.
The evidence adapter combines each draw with the current belief to obtain
$p_t^{(s)}(k\mid a)$. Let $O_{ij}=\operatorname{IoU}(M_i,M_j)$,
$\hat k^{(s)}=\arg\max_k p_t^{(s)}(k\mid a)$, and
$\hat k_t=\arg\max_k p_t(k)$. We measure the expected improvement in the
posterior-weighted overlap of the selected mask with the candidate bank:
{\small \setlength\abovedisplayskip{2pt}\setlength\belowdisplayskip{2pt}
\begin{equation}
\begin{aligned}
 \widehat U_t(a)
 ={}&\frac{1}{S}\sum_{s=1}^{S}\sum_k
       p_t^{(s)}(k\mid a)\,O_{\hat k^{(s)}k}
    -\sum_k p_t(k)\,O_{\hat k_t k}.
\end{aligned}
\label{eq:view_utility}
\end{equation}}
Both terms use the MAP mask decision in Eq.~\eqref{eq:mask_output}.
Using candidate-mask overlap allows NBE to distinguish between changes among
similar mask extents and changes between spatially distinct objects. Importantly,
this is computed entirely from predicted observations and candidate
masks; the unknown target mask is never used during planning.

\input{tables/active_listening}

We use common random draws across candidate actions to reduce pose-comparison
noise. The cross-entropy method searches over navigable ground-plane positions
and yaw and selects the pose with the highest predicted utility. Only the
selected pose is executed and recorded; imagined observations are used solely
for planning and never update the retained belief.

\vspace{2pt}
\noindent\textbf{Feedback for Subsequent Listening.}
After moving to the selected pose, the acquired recording updates both the
object--site belief and the acoustic memory used for subsequent planning.
Let $\lambda_{kh}$ denote the masked component score of the new observation,
$c_k=\log\sum_h\rho_t(h\mid k)e^{\lambda_{kh}}$, and $d_k$ the
candidate-level score increment produced by the evidence adapter, excluding
the current prior. The joint belief is updated as
{\small \setlength\abovedisplayskip{2pt}\setlength\belowdisplayskip{2pt}
\begin{equation}
\begin{aligned}
 \Lambda_{kh}=\lambda_{kh}+d_k-c_k;
 \quad q_{t+1}(k,h)=
 \frac{q_t(k,h)e^{\Lambda_{kh}}}
 {\sum_{i,j}q_t(i,j)e^{\Lambda_{ij}}}.
\end{aligned}
\label{eq:belief_update}
\end{equation}}
This correction preserves the calibrated object-level evidence while retaining
the conditional belief over emission sites. When no neural score correction is
used, $d_k=c_k$, and the update reduces to multiplication by the component
likelihood scores, ensuring that the initial prior is incorporated only once.
The acquired recording also updates the acoustic state used to predict future
observations. At the selected pose, inactive covariance bands are replaced by
their predicted values before assimilation. The frozen recurrent cell produces
$\mathbf z^{\mathrm{assim}}_{kh}$, after which a shared channel-wise gate updates
the memory:
{\setlength\abovedisplayskip{2pt}\setlength\belowdisplayskip{2pt}
\begin{equation}
 \mathbf z_{kh}^{(t+1)}=\mathbf z^-_{kh}
 +2\sigma(\mathbf g)\odot
  (\mathbf z^{\mathrm{assim}}_{kh}-\mathbf z^-_{kh}).
\label{eq:memory_update}
\end{equation}}
The shared gate contains 128 parameters and is initialized to recover the
original assimilation update. If a recording contains no usable acoustic
evidence, the belief remains unchanged while only the predicted state is
propagated.

The updated belief determines which source hypotheses are emphasized in the
next NBE step, while the updated memories determine their future acoustic
predictions. This closes the listening loop, allowing each subsequent action
to exploit evidence collected at earlier poses. Once the sensing budget is
exhausted, Eq.~\eqref{eq:mask_output} selects the final mask on the reference
LiDAR cloud.


%% file: tables/active_listening.tex
\begin{table*}[t!]
\centering
\caption{\textbf{Active audio-prompted LiDAR segmentation on AudLiSeg-Indoor.}
Comparison across Near-Field, Mid-Field, and Far-Field
settings under multiple ($>1$) Listening Positions (LPs, $T+1s$).
Overall results are macro-averaged across the three settings.
We report mean mask IoU (mIoU), F-score (F1),
exact source identification accuracy (Acc.),
and cumulative receiver travel (Dist.).
}
\label{tab:active_main}
\vspace{-5pt}
\setlength{\tabcolsep}{5pt}
\renewcommand{\arraystretch}{1.12}

\resizebox{\textwidth}{!}{%
\begin{tabular}{@{}l c
*{4}{>{\columncolor{NearColor}}c}
*{4}{>{\columncolor{MidColor}}c}
*{4}{>{\columncolor{FarColor}}c}
*{4}{c}@{}}

\toprule

& &
\multicolumn{4}{>{\columncolor{NearColor}}c}
{\textbf{Near-Field}} &
\multicolumn{4}{>{\columncolor{MidColor}}c}
{\textbf{Mid-Field}} &
\multicolumn{4}{>{\columncolor{FarColor}}c}
{\textbf{Far-Field}} &
\multicolumn{4}{c}
{\textbf{Overall}} \\

\cmidrule(lr){3-6}
\cmidrule(lr){7-10}
\cmidrule(lr){11-14}
\cmidrule(l){15-18}

\textbf{Method} &
\textbf{LPs} &
\textbf{mIoU $\uparrow$} &
\textbf{F1 $\uparrow$} &
\textbf{Acc. $\uparrow$} &
\textbf{Dist. $\downarrow$} &
\textbf{mIoU $\uparrow$} &
\textbf{F1 $\uparrow$} &
\textbf{Acc. $\uparrow$} &
\textbf{Dist. $\downarrow$} &
\textbf{mIoU $\uparrow$} &
\textbf{F1 $\uparrow$} &
\textbf{Acc. $\uparrow$} &
\textbf{Dist. $\downarrow$} &
\textbf{mIoU $\uparrow$} &
\textbf{F1 $\uparrow$} &
\textbf{Acc. $\uparrow$} &
\textbf{Dist. $\downarrow$} \\

\midrule


\multirow{4}{*}{\textbf{ActListen (Ours)}}
& 2 & \textbf{44.22} & \textbf{52.79} & \textbf{64.60} & 1.69
    & \textbf{42.18} & \textbf{50.23} & \textbf{57.87} & 2.24
    & \textbf{43.23} & \textbf{47.85} & \textbf{57.14} & 2.07
    & \textbf{43.21} & \textbf{50.29} & \textbf{59.87} & 2.00 \\

& 3 & \textbf{45.14} & \textbf{54.05} & 64.83 & 2.57
    & 43.55 & 51.86 & 60.43 & 3.18
    & 43.81 & 48.41 & 56.12 & 3.11
    & 44.17 & 51.44 & 60.46 & 2.95 \\

& 4 & \textbf{45.40} & \textbf{54.70} & \textbf{65.29} & 3.41
    & \textbf{45.44} & \textbf{54.11} & \textbf{63.83} & 4.12
    & 42.80 & 47.35 & 53.06 & 4.93
    & \textbf{44.55} & \textbf{52.05} & \textbf{60.73} & 4.15 \\

& 5 & 45.40 & 54.71 & \textbf{65.06} & 4.19
    & \textbf{45.84} & \textbf{54.51} & 63.83 & 5.12
    & \textbf{45.15} & \textbf{50.13} & \textbf{58.16} & 5.52
    & \textbf{45.46} & \textbf{53.12} & \textbf{62.35} & 4.94 \\

\midrule


\multirow{4}{*}{Prior-Forecast}
& 2 & 43.68 & 52.01 & 64.37 & 1.60
    & 41.33 & 48.98 & 56.60 & 2.12
    & 39.87 & 44.29 & 52.04 & 2.08
    & 41.63 & 48.43 & 57.67 & 1.93 \\

& 3 & 44.42 & 53.20 & 63.91 & 2.56
    & 43.21 & 51.39 & 59.57 & 3.24
    & 42.23 & 46.81 & 55.10 & 3.27
    & 43.29 & 50.47 & 59.53 & 3.02 \\

& 4 & 44.84 & 53.73 & 64.14 & 3.48
    & 44.10 & 52.45 & 60.43 & 4.30
    & 42.61 & 47.37 & \textbf{56.12} & 4.50
    & 43.85 & 51.18 & 60.23 & 4.09 \\

& 5 & \textbf{45.67} & \textbf{54.86} & 64.60 & 4.40
    & 44.70 & 53.19 & 61.70 & 5.42
    & 43.05 & 47.95 & \textbf{58.16} & 5.62
    & 44.47 & 52.00 & 61.49 & 5.15 \\

\midrule


\multirow{4}{*}{DirectValue}
& 2 & 43.95 & 52.51 & \textbf{64.60} & 1.94
    & 40.82 & 48.47 & 57.02 & 2.53
    & 41.62 & 46.00 & 53.06 & 2.51
    & 42.13 & 48.99 & 58.23 & 2.33 \\

& 3 & 44.01 & 52.85 & 64.14 & 2.95
    & \textbf{43.76} & \textbf{52.10} & 60.85 & 3.81
    & \textbf{44.97} & \textbf{49.60} & \textbf{57.14} & 4.52
    & \textbf{44.25} & \textbf{51.52} & \textbf{60.71} & 3.76 \\

& 4 & 44.43 & 53.51 & 63.68 & 4.01
    & 44.30 & 52.85 & 62.98 & 5.27
    & \textbf{44.47} & \textbf{49.15} & 55.10 & 6.61
    & 44.40 & 51.84 & 60.59 & 5.30 \\

& 5 & 45.01 & 54.24 & 63.68 & 5.11
    & 45.36 & 54.10 & \textbf{65.53} & 6.73
    & 44.30 & 49.15 & 57.14 & 8.64
    & 44.89 & 52.50 & 62.12 & 6.83 \\

\midrule


\multirow{4}{*}{Analytical Planner}
& 2 & 43.54 & 51.74 & 63.22 & 2.69
    & 41.11 & 48.70 & 56.60 & 2.85
    & 40.63 & 44.84 & 51.02 & 2.88
    & 41.76 & 48.43 & 56.95 & 2.81 \\

& 3 & 45.11 & 53.59 & \textbf{65.06} & 5.45
    & 42.25 & 50.22 & 59.15 & 5.61
    & 43.22 & 47.68 & 54.08 & 5.76
    & 43.53 & 50.50 & 59.43 & 5.61 \\

& 4 & 44.33 & 53.02 & 64.14 & 8.09
    & 42.68 & 50.98 & 60.00 & 8.29
    & 42.29 & 46.68 & 54.08 & 8.34
    & 43.10 & 50.23 & 59.41 & 8.24 \\

& 5 & 43.95 & 52.67 & 63.22 & 10.80
    & 43.25 & 51.79 & 62.98 & 11.00
    & 44.23 & 48.72 & 55.10 & 11.41
    & 43.81 & 51.06 & 60.43 & 11.07 \\

\midrule


\multirow{4}{*}{Random Pose}
& 2 & 42.16 & 49.87 & 59.31 & 1.72
    & 39.57 & 47.01 & 54.47 & 1.87
    & 38.04 & 42.03 & 47.96 & 1.64
    & 39.92 & 46.30 & 53.91 & 1.74 \\

& 3 & 42.94 & 51.09 & 61.38 & 3.42
    & 40.04 & 47.71 & 55.74 & 3.45
    & 40.39 & 44.71 & 53.06 & 3.24
    & 41.12 & 47.84 & 56.73 & 3.37 \\

& 4 & 42.52 & 50.88 & 60.92 & 5.04
    & 39.57 & 47.16 & 54.89 & 5.18
    & 40.46 & 44.76 & 53.06 & 8.57
    & 40.85 & 47.60 & 56.29 & 6.26 \\

& 5 & 42.62 & 51.08 & 60.69 & 6.65
    & 40.46 & 48.10 & 56.17 & 6.77
    & 41.18 & 45.54 & 54.08 & 10.51
    & 41.42 & 48.24 & 56.98 & 7.98 \\

\bottomrule
\end{tabular}%
}
\end{table*}

%% file: sections/4_experiments.tex
\section{Experiments and Evaluation}
\label{sec:experiments}
\vspace{-3pt}

ActListen is trained in stages while keeping the perception frontend and base acoustic model frozen. We first train the evidence adapter using cross-entropy with soft candidate targets proportional to $\operatorname{IoU}(M_k,M^*)^4$. With the adapter fixed, the observation model is optimized using a predictive likelihood objective, a KL consistency term between adapter outputs on real and imagined observations, and a penalty on the mean correction. The memory gate is trained separately on sequences of measured covariances using a reliability-weighted future prediction loss. Ground-truth labels are used only during training and validation to select source hypotheses for gate supervision and are not available at inference. Pose selection itself is not trained by differentiating through the final segmentation objective.

\begin{figure*}[t!]
  \centering
  \includegraphics[width=0.98\linewidth]{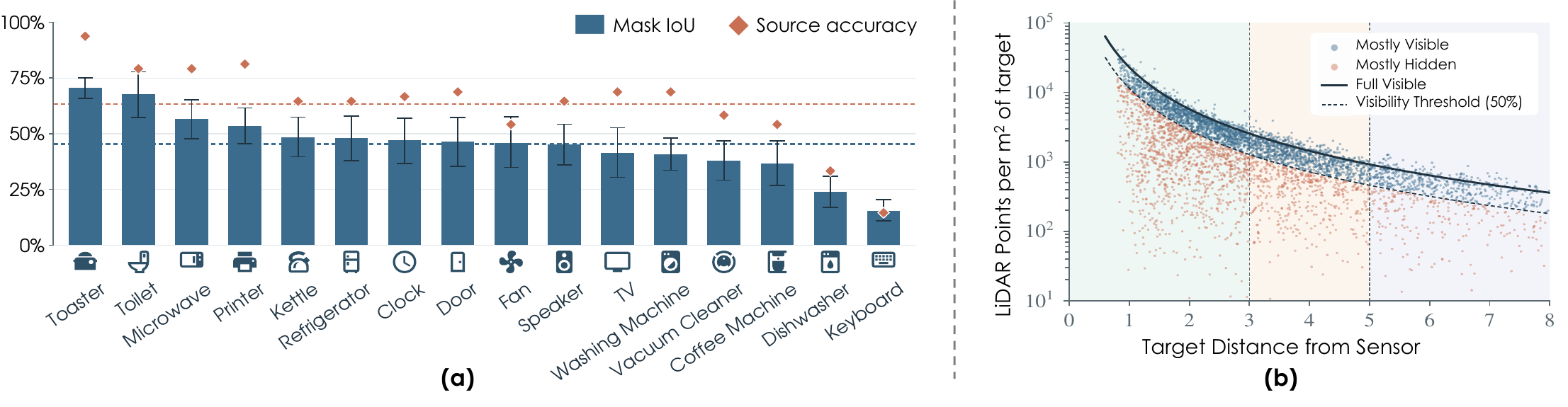}
\vspace{-14pt}
\caption{\textbf{(a)} Per-class ActListen performance on AudLiSeg-Indoor with five listening positions, where \textcolor{blue}{-\,-\,-} and \textbf{\textcolor{orange}{-\,-\,-}} indicate the means of mask IoU and source-accuracy, respectively.
\textbf{(b)} LiDAR point density distribution across object observations in AudLiSeg-Indoor.
}
\label{fig:performance_visibility}
\end{figure*}

\subsection{ActListen Evaluation on AudLiSeg}
\label{sec:audliseg_evaluation}
\vspace{-3pt}

\noindent\textbf{Indoor Evaluation.}
From a single LiDAR scan and recording, ActListen identifies and segments sounding objects at 39.49\% mIoU on the unseen test set (Table~\ref{tab:single_view_ablation}). 

Sequential acoustic acquisition further improves segmentation. With five listening positions, ActListen achieves \textbf{45.40\%}, \textbf{45.84\%}, and \textbf{45.15\%} mIoU in the Near-Field, Mid-Field, and Far-Field settings, respectively (Table~\ref{tab:active_main}), corresponding to relative gains of \textbf{6.52\%}, \textbf{13.30\%}, and \textbf{9.64\%} over Random Pose. Fig.~\ref{fig:teaser}\textcolor{cvprblue}{c} illustrates how additional listening positions recover initially missed or partially segmented objects, demonstrating the benefit of active listening in resolving acoustic ambiguity. Fig.~\ref{fig:performance_visibility}\textcolor{cvprblue}{a} further shows class-wise performance, highlighting variation across object categories due to differences in acoustic characteristics and scene geometry.

Fig.~\ref{fig:actlisten_qual}\textcolor{cvprblue}{a} and
Fig.~\ref{fig:actlisten_qual}\textcolor{cvprblue}{b} show qualitative indoor results of ActListen. Results shows that additional recordings progressively improve source identification and mask completeness from an initially partial prediction, while operating on the same reference LiDAR scan without acquiring additional geometry.
Fig.~\ref{fig:actlisten_qual}\textcolor{cvprblue}{c} further presents results across different distractor states, illustrating ActListen's robustness as acoustic ambiguity increases.

\vspace{2pt}
\noindent\textbf{Outdoor Evaluation.}
ActListen uses real LiDAR geometry and engine, horn,
and siren sounds (Table~\ref{tab:outdoor_results}). With two listening positions,
Near-Field mIoU peaks at \textbf{53.86\%}, while Mid-Field reaches
\textbf{22.00\%}. In the harder Far-Field setting, one to five positions
raise mIoU from \textbf{7.95\%} to \textbf{22.97\%} and F1 from
\textbf{12.11\%} to \textbf{29.81\%}.

\vspace{-5pt}
\begin{tcolorbox}[
  colback=white!5,
  colframe=black!50,
  boxrule=1pt,
  arc=1pt,
  left=2pt,
  right=2pt,
  top=2pt,
  bottom=2pt,
  width=\linewidth,
  breakable,
  enhanced
]
The indoor and outdoor evaluations demonstrate that active acoustic evidence acquisition can improve sound-source segmentation across different environments, with particularly substantial gains for distant outdoor sources.
\end{tcolorbox}
\vspace{-5pt}

\input{tables/outdoor}

\subsection{Comparison with Active Listening Strategies}
\label{sec:comparision_active_listening}

We compare ActListen with four alternatives under identical LiDAR, audio, candidate masks, evidence processing, navigability, and observation budgets ($T+1\in\{2,3,4,5\}$). \textbf{Prior-Forecast} uses the same planning objective and assimilation as ActListen but removes the learned forecast correction. \textbf{DirectValue} predicts action values directly from geometry, acoustic statistics, beliefs, and latent context without imagining future observations. \textbf{Analytical Planner} selects poses using belief-weighted geometric bearing changes, while \textbf{Random Pose} samples feasible positions without evaluating acoustic informativeness. Later views add only audio and receiver poses, not LiDAR. 
\textit{Additional details are in Suppl. Mat. Sec~\ref{supp_sec:baseline_details}}.

Across the three distance tiers and four observation budgets in Table~\ref{tab:active_main}, ActListen achieves the highest mIoU in \textbf{8 of 12} settings. From two to five views, its mIoU improves by \textbf{1.18}, \textbf{3.66}, and \textbf{1.92} points in Near-, Mid-, and Far-Field, respectively, although gains are not strictly monotonic. At five views, ActListen also yields the lowest cumulative travel across all three tiers.


\subsection{Comparison with Existing Methods}
\label{sec:comparision_existing_methods}

\subsubsection{Comparison with EchoSegnet}

To our knowledge, no existing method directly addresses audio-prompted LiDAR segmentation. The closest 3D audiovisual method, \textbf{EchoSegnet}~\cite{sokolov20243d}, grounds sounding objects using RGB, binaural audio, and reconstructed 3D Gaussians rather than LiDAR. To examine whether ActListen's acoustic reasoning transfers beyond LiDAR, we integrate its acoustic evidence into EchoSegnet's proposal-selection pipeline while retaining the pretrained CAOD, SAM, ImageBind, and 3D reconstruction components.

\noindent\textbf{Experimental Setup.}
EchoSegnet provides object proposals and audiovisual similarity scores. For each proposal, its 3D Gaussian centers are used to aggregate ActListen's acoustic predictions, sequential beliefs, and recurrent world-model features into a 26-dimensional representation. These features are combined with ImageBind similarity and proposal statistics, and a lightweight two-layer MLP predicts a residual proposal score. All pretrained encoders and the acoustic world model remain frozen.

We evaluate both methods on the 34 scenes of 3DAVS-S34-O7 across four buildings using building-disjoint splits. Both methods use the same RGB-audio inputs and 3D Gaussian reconstruction. Following EchoSegnet's protocol, 90 held-out views are evaluated using mask IoU, visible-target IoU, and $F$-measure. 
\textit{Additional implementation details are provided in Suppl. Mat. Sec.~\ref{supp_sec:echosegnet}}.

\begin{figure}[t!]
  \centering
  \includegraphics[width=1.0\linewidth]{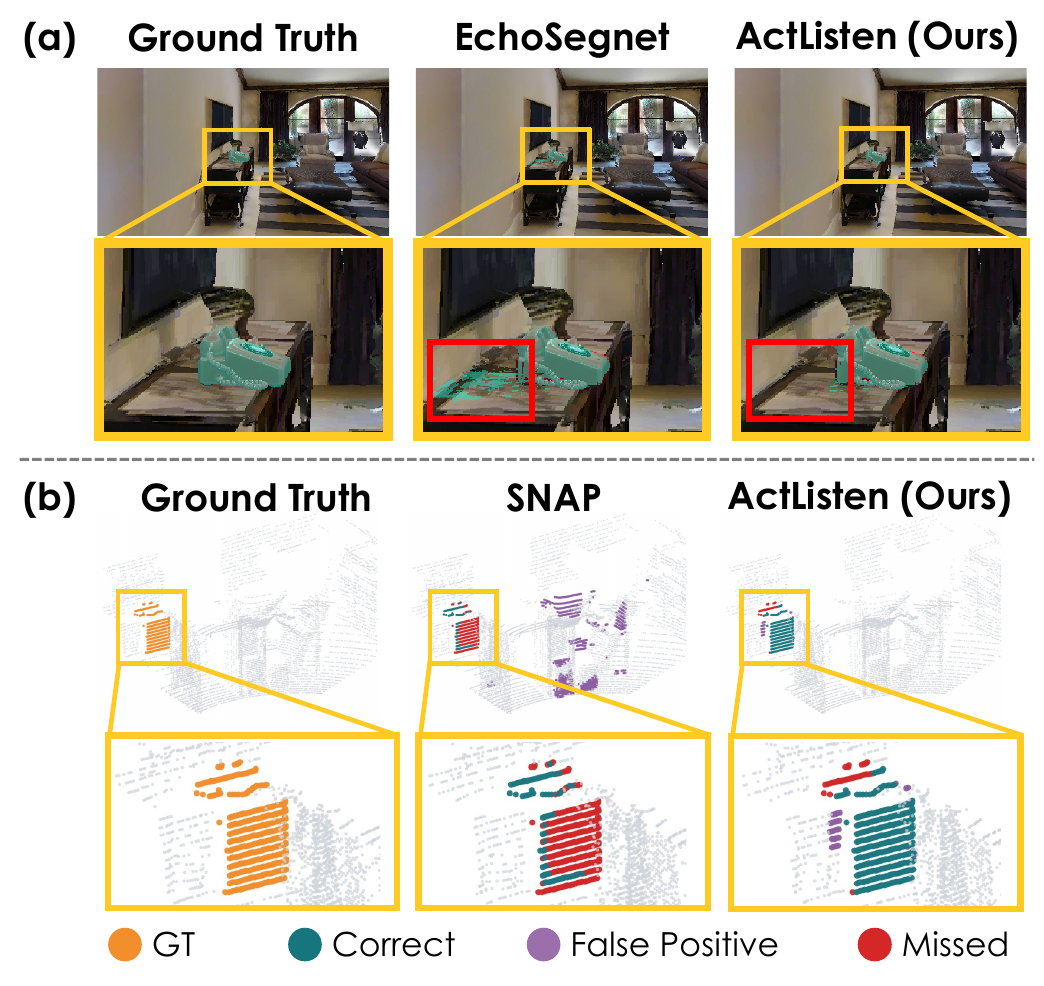}
  \vspace{-20pt}
  \caption{\textbf{Qualitative comparisons} of \textbf{(a)} EchoSegnet and ActListen on audiovisual 3D segmentation. \textbf{(b)} SNAP + GT Category, prompted with ``visible washing machine objects'', and ActListen on audio-prompted LiDAR segmentation.}
  \label{fig:echosegnet_snap_qual}
  \vspace{-5pt}
\end{figure}

\noindent\textbf{Results.}
Integrating ActListen's acoustic evidence improves mask IoU, visible-target IoU, and $F$-measure over EchoSegnet by \textbf{7.34}, \textbf{12.47}, and \textbf{15.00} percentage points, respectively (Table~\ref{tab:echosegnet_adaptation}). These gains show that ActListen's acoustic representations transfer beyond LiDAR grounding and improve 3D audiovisual segmentation under the same reconstruction pipeline. Fig.~\ref{fig:echosegnet_snap_qual}\textcolor{cvprblue}{a} illustrates qualitative comparisons between EchoSegnet and ActListen.

\subsubsection{Comparison with SNAP}
SNAP~\cite{gupta2026snap} supports language-prompted point-cloud segmentation but not audio. We adapt it using two prompting strategies: \textit{CLAP-predicted categories}, representing an audio-to-language pipeline, and \textit{ground-truth categories}, providing oracle semantic information to isolate category-recognition errors.

\noindent\textbf{Experimental Setup.}
We fine-tune the SNAP-C decoder on AudLiSeg-Indoor LiDAR clouds and category masks. At inference, frozen CLAP predicts one of 16 sound categories, which CLIP encodes as SNAP's language prompt; the oracle variant instead uses the ground-truth category. We evaluate on S1 and S2, where each reference contains a single target instance, avoiding same-category instance ambiguity. Both methods use identical LiDAR scans, with ActListen additionally receiving four-channel audio.

\noindent\textbf{Results.}
ActListen outperforms both SNAP variants across all metrics on S1 and S2 (Table~\ref{tab:snap_comparison}, Fig.~\ref{fig:echosegnet_snap_qual}\textcolor{cvprblue}{b}). On S2, ActListen exceeds SNAP with oracle and CLAP-predicted categories by \textbf{15.23} and \textbf{23.09} mIoU, respectively. The improvement over oracle prompts demonstrates the benefit of directly grounding acoustic evidence in LiDAR object instances beyond semantic category recognition.
\textit{Additional details are provided in Suppl. Mat. Sec.~D.5}.

\vspace{-5pt}
\begin{tcolorbox}[
  colback=white!5,
  colframe=black!50,
  boxrule=1pt,
  arc=1pt,
  left=2pt,
  right=2pt,
  top=2pt,
  bottom=2pt,
  width=\linewidth,
  breakable,
  enhanced
]
Together, the results in Sec.~\ref{sec:comparision_active_listening} and Sec.~\ref{sec:comparision_existing_methods} demonstrate ActListen's effective world-model-guided exploration, superior active listening, and transferability to audiovisual 3D segmentation and LiDAR grounding.
\end{tcolorbox}
\vspace{-5pt}




\input{tables/echosegnet}
\input{tables/snap}

\input{tables/singleview}

\subsection{Ablations}
\vspace{-3pt}
We ablate the single-view grounding components by removing (i) predicted-observed covariance agreement (\textit{w/o Spatial Acoustic Evidence}), (ii) Point-Bind/CLAP sound--object compatibility (\textit{w/o Semantic Evidence}), and (iii) both while retaining the audio-conditioned frontend scores (\textit{Selector Prior Only}). All variants use identical inputs and frozen components.

As shown in Table~\ref{tab:single_view_ablation}, full ActListen achieves the highest mIoU, F1, recall, and source accuracy. Removing spatial acoustic evidence, semantic evidence, or both decreases mIoU by \textbf{1.06}, \textbf{0.95}, and \textbf{1.77} points, respectively. Scene-cluster bootstrap confidence intervals exclude zero for all three drops, confirming that spatial acoustic and semantic cues provide complementary grounding evidence.

%% file: tables/outdoor.tex
\begin{table}[t!]
\centering
\caption{Evaluation of ActListen on AudLiSeg-Outdoor across increasing numbers of listening positions (LPs).}
\label{tab:outdoor_results}
\vspace{-4pt}
\setlength{\tabcolsep}{4pt}
\renewcommand{\arraystretch}{1.12}
\resizebox{\columnwidth}{!}{%
\begin{tabular}{@{}c
*{3}{>{\columncolor{NearColor}}c}
*{3}{>{\columncolor{MidColor}}c}
*{3}{>{\columncolor{FarColor}}c}@{}}
\toprule
&
\multicolumn{3}{>{\columncolor{NearColor}}c}{\textbf{Near-Field}} &
\multicolumn{3}{>{\columncolor{MidColor}}c}{\textbf{Mid-Field}} &
\multicolumn{3}{>{\columncolor{FarColor}}c}{\textbf{Far-Field}} \\
\cmidrule(lr){2-4}
\cmidrule(lr){5-7}
\cmidrule(l){8-10}

\textbf{LPs} &
\textbf{mIoU $\uparrow$} &
\textbf{F1 $\uparrow$} &
\textbf{Acc. $\uparrow$} &
\textbf{mIoU $\uparrow$} &
\textbf{F1 $\uparrow$} &
\textbf{Acc. $\uparrow$} &
\textbf{mIoU $\uparrow$} &
\textbf{F1 $\uparrow$} &
\textbf{Acc. $\uparrow$} \\
\midrule

1 & 51.88 & 62.35 & 88.68
  & 19.17 & 29.12 & 84.62
  & 7.95 & 12.11 & 56.88 \\

2 & \textbf{53.86} & \textbf{64.42} & \textbf{90.57}
  & \textbf{22.00} & 31.40 & 83.65
  & 20.71 & 26.86 & 74.31 \\

3 & 52.72 & 62.90 & 88.68
  & 21.98 & \textbf{31.46} & \textbf{85.58}
  & 22.27 & 29.09 & \textbf{76.15} \\

4 & 52.82 & 62.92 & 88.68
  & 19.92 & 28.82 & 76.92
  & 20.86 & 27.32 & 74.31 \\

5 & 52.47 & 62.31 & 88.68
  & 19.84 & 28.67 & 75.00
  & \textbf{22.97} & \textbf{29.81} & \textbf{76.15} \\

\bottomrule
\end{tabular}%
}
\vspace{-15pt}
\end{table}

%% file: tables/echosegnet.tex
\begin{table}[t!]
\centering
\caption{\textbf{Comparison with EchoSegnet}~\cite{sokolov20243d} on 3DAVS-S34-O7.}
\label{tab:echosegnet_adaptation}
\vspace{-8pt}
\setlength{\tabcolsep}{3pt}
\renewcommand{\arraystretch}{1.12}
\resizebox{\columnwidth}{!}{%
\begin{tabular}{@{}lccc@{}}
\toprule
\textbf{Method} &
\textbf{Mask IoU $\uparrow$} &
\textbf{Visible IoU $\uparrow$} &
\textbf{F-Score $\uparrow$} \\
\midrule
EchoSegnet
& 82.11 & 69.63 & 74.92 \\

\rowcolor{blue!8} \textbf{ActListen Adaptation (Ours)}
& \textbf{89.46}
& \textbf{82.10}
& \textbf{89.92} \\
\bottomrule
\end{tabular}%
}
\end{table}

%% file: tables/snap.tex
\begin{table}[t!]
\centering
\caption{\textbf{Comparison with SNAP}~\cite{gupta2026snap} on AudLiSeg}
\label{tab:snap_comparison}
\vspace{-8pt}
\setlength{\tabcolsep}{5pt}
\renewcommand{\arraystretch}{1.12}
\resizebox{\columnwidth}{!}{%
\begin{tabular}{@{}l l c c c@{}}
\toprule
\textbf{Tier} & \textbf{Method}
& \textbf{Mask IoU} $\uparrow$
& \textbf{F1} $\uparrow$
& \textbf{IoU $>0.5$} $\uparrow$ \\
\midrule

\multirow{3}{*}{\shortstack[l]{S1}}
& SNAP + GT Category
& 43.76 & 52.95 & 41.67 \\
& SNAP + CLAP Category
& 40.34 & 48.87 & 37.50 \\
& \textbf{ActListen (Single-View)}
& \textbf{60.97} & \textbf{67.04} & \textbf{62.50} \\

\midrule

\multirow{3}{*}{\shortstack[l]{S2}}
& SNAP + GT Category
& 24.41 & 30.22 & 22.35 \\
& SNAP + CLAP Category
& 16.55 & 20.29 & 15.89 \\
& \textbf{ActListen (Single-View)}
& \textbf{39.64} & \textbf{46.19} & \textbf{43.87} \\

\bottomrule
\end{tabular}%
}
\end{table}

%% file: tables/singleview.tex
\begin{table}[t!]
\centering
\caption{\textbf{Ablation study of single-view ActListen on AudLiSeg-Indoor.} Evaluating the contributions of spatial acoustic and semantic evidence to sound-source segmentation.}
\label{tab:single_view_ablation}
\vspace{-8pt}
\setlength{\tabcolsep}{4pt}
\renewcommand{\arraystretch}{1.12}
\resizebox{\columnwidth}{!}{%
\begin{tabular}{@{}lccccc@{}}
\toprule
\textbf{Configuration} &
\textbf{mIoU $\uparrow$} &
\textbf{F1 $\uparrow$} &
\textbf{Prec. $\uparrow$} &
\textbf{Recall $\uparrow$} &
\textbf{Acc. $\uparrow$} \\
\midrule

Selector prior only
& 37.73 & 44.27 & 46.55 & 49.10 & 52.34 \\

w/o Spatial acoustic evidence
& 38.43 & 44.96 & 45.79 & 51.38 & 52.73 \\

w/o Semantic evidence
& 38.55 & 45.25 & \textbf{47.64} & 50.08 & 53.52 \\

\midrule
\rowcolor{blue!8} \textbf{ActListen (Full)}
& \textbf{39.49}
& \textbf{46.20}
& 47.28
& \textbf{52.60}
& \textbf{54.04} \\

\bottomrule
\end{tabular}%
}
\vspace{-15pt}
\end{table}

%% file: sections/5_conclusion.tex
\section{Conclusion}
\label{sec:conclusion}
\vspace{-5pt}
In this work, we introduced \textbf{AudLiSeg}, a benchmark for audio-prompted LiDAR segmentation, and \textbf{ActListen}, an active framework that couples hypothesis initialization with closed-loop predictive listening. By forecasting acoustic observations and updating source beliefs from acquired recordings, ActListen improves sounding-object segmentation without RGB or additional LiDAR scans. 
Experiments show that active listening improves segmentation performance over single-view, while the learned acoustic representations also transfer to 3D audiovisual segmentation.\\ 
\textbf{Future Work.} Future work will extend ActListen to dynamic acoustic scenes, moving sound sources, and real-world multimodal sensing.

%% file: suppl-main.tex
\definecolor{AudNear}{HTML}{EFF7F4}
\definecolor{AudMid}{HTML}{FCF5EE}
\definecolor{AudFar}{HTML}{F3F4FA}

\section{Overview}
\label{supp_sec:overview}

This supplementary material provides additional details on the
\textbf{AudLiSeg} benchmark, the \textbf{ActListen} framework,
and the experimental evaluation.

\begin{tcolorbox}[
  colback=blue!3,
  colframe=blue!3,
  boxrule=0pt,
  arc=3pt,
  left=3pt,
  right=3pt,
  top=3pt,
  bottom=3pt,
  width=\linewidth,
  breakable,
  enhanced
]
\begin{itemize}
    \item \textbf{Sec.~\ref{supp_sec:aduliseg}} describes the construction of
    AudLiSeg-Indoor and AudLiSeg-Outdoor, including spatial audio
    generation, dataset statistics, distance-based difficulty,
    source distributions, and instance ambiguity settings.
    The \textit{\textbf{supplementary video}} provides representative
    examples from both datasets, illustrating the LiDAR scenes,
    sound-producing objects, and corresponding audio recordings.

    \item \textbf{Sec.~\ref{supp_sec:actlisten}} presents the implementation
    details of ActListen, covering acoustic representations,
    geometry-conditioned world modeling, belief updates,
    Next-Best-Ear planning, and training configurations.

    \item \textbf{Sec.~\ref{supp_sec:evaluation}} details the evaluation metrics,
    active listening protocol, baseline implementations, and
    adaptations of existing methods.

    \item \textbf{Sec.~\ref{supp_sec:additional_results}} presents additional
    qualitative results. \textbf{Sec.~\ref{supp_sec:limitations}} discusses the limitations
    and failure cases.
\end{itemize}
\end{tcolorbox}

\section{AudLiSeg Benchmark Details}
\label{supp_sec:aduliseg}

\paragraph{Task and Information Boundary.}
AudLiSeg evaluates audio-prompted 3D LiDAR instance segmentation, where the objective is to identify and segment the physical object producing sound. Given a reference LiDAR point cloud
$\mathcal{P}=\{\mathbf{x}_n\}_{n=1}^{N}$ and a synchronized four-channel audio recording
$\mathbf{A}_0\in\mathbb{R}^{88{,}200\times4}$ acquired at receiver pose $\mathbf{p}_0$, the task is to predict the point-level mask of the active source instance. Each recording contains two seconds of audio sampled at $44.1$~kHz.

Each observation contains exactly one active sound-producing object, although other sound-capable objects may be present in the scene. Consequently, the target is the specific emitting instance rather than all objects belonging to the same semantic category. Both the indoor and the outdoor branches provide human-verified point-level active-source annotations and employ a calibrated tetrahedral microphone array.

ActListen operates without RGB input. During active sensing, additional recordings are acquired at selected receiver poses, while the reference LiDAR cloud remains fixed. Ground-truth source locations, instance identities, target masks, and simulated impulse responses are used for dataset construction, training, or evaluation, but are not provided as ground-truth information during inference.

\subsection{Indoor Benchmark Construction}
\label{supp_sec:indoor_construction}

AudLiSeg-Indoor is constructed from semantically annotated Habitat--Matterport 3D (HM3D) scenes~\cite{ramakrishnan2021habitat}. We simulate $64$-beam, $360^\circ$ LiDAR scans from the semantic meshes, using $1{,}800$ azimuthal samples per scan. Ray--mesh intersections preserve the corresponding object-instance annotations, enabling point-level supervision of the sounding object. Acoustic propagation is simulated separately using material-dependent properties.

The benchmark includes 16 sound-capable object categories: refrigerator, fan, television, door, toilet, washing machine, vacuum cleaner, clock, microwave, dishwasher, toaster, coffee machine, speaker, printer, keyboard, and kettle. Each source corresponds to a fixed physical object instance observed from 12 navigable receiver poses. In total, AudLiSeg-Indoor contains \textbf{4,800 observations} from \textbf{400 source instances} across \textbf{52 scenes}. 

\begin{figure}[h]
    \centering
    \includegraphics[width=1.0\columnwidth]{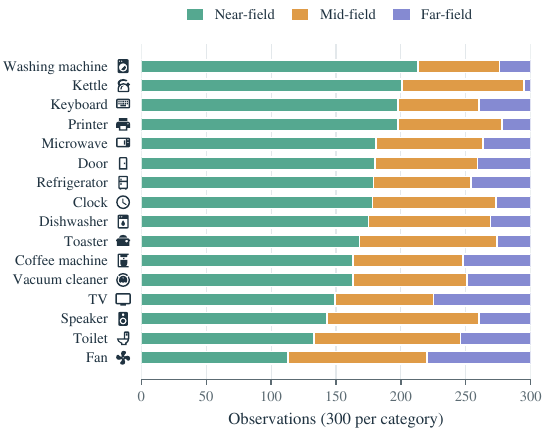}
    \vspace{-15pt}
    \caption{\textbf{Category-wise distribution of AudLiSeg-Indoor.}
    Each of the 16 object categories contains 300
    observations, partitioned into Near-, Mid-,
    and Far-Field distance tiers.}
    \label{suppl_fig:indoor_categories}
    \vspace{-15pt}
\end{figure}

The category-wise distribution is illustrated in
Fig.~\ref{suppl_fig:indoor_categories}. Each of the
16 categories contributes 300 observations, ensuring
balanced semantic coverage, while the proportions
of Near-, Mid-, and Far-Field observations vary
across object categories.

\paragraph{Spatial Audio Generation.}
A four-microphone tetrahedral array is centered at the LiDAR origin, providing spatial acoustic observations from a shared receiver coordinate frame. In the local $(x\text{-forward},y\text{-left},z\text{-up})$ coordinate system, the microphone offsets are
\begin{equation}
\begin{aligned}
\boldsymbol{\delta}_a &= u(1,1,1), &
\boldsymbol{\delta}_b &= u(1,-1,-1),\\
\boldsymbol{\delta}_c &= u(-1,1,-1), &
\boldsymbol{\delta}_d &= u(-1,-1,1),
\end{aligned}
\label{eq:audliseg_tetrahedron}
\end{equation}
where $u=0.03266$~m, corresponding to a pairwise microphone spacing of approximately $9.24$~cm.

We employ SoundSpaces~2.0~\cite{chen2020soundspaces,chen2022soundspaces} to simulate acoustic propagation, including direct sound, reflections, diffraction, and transmission under material-dependent acoustic properties. For each source--receiver configuration, the source waveform is convolved with the corresponding room impulse response at each microphone. The resulting recordings are synchronized, two-second, four-channel waveforms sampled at $44.1$~kHz. The shared source recording and spatially varying impulse responses preserve inter-channel acoustic cues.

\subsection{Outdoor Benchmark Construction}
\label{supp_sec:outdoor_construction}

AudLiSeg-Outdoor combines real-world LiDAR geometry from KITTI HDL-64E scans with SemanticKITTI point-level instance annotations~\cite{behley2019semantickitti} and sensor poses. Unlike the simulated indoor geometry, the outdoor branch uses real driving-scene point clouds while generating spatial audio through an analytical acoustic propagation model.

Engine, horn, and siren recordings from
ESC-50~\cite{piczak2015esc} are assigned to
annotated vehicle locations spanning four
SemanticKITTI classes: Car, Moving Car, Truck,
and Other Vehicle. Car and Moving Car are
distinguished by their original motion annotations,
with the latter indicating vehicles observed
moving during the recorded sequence.
Acoustic propagation incorporates a direct sound path and a planar ground-reflection component. The resulting impulse responses are evaluated at 12 receiver poses using the same four-microphone tetrahedral configuration as the indoor branch. Each microphone signal is synthesized by convolving a two-second source segment with its corresponding impulse response, producing synchronized four-channel audio at $44.1$~kHz.

AudLiSeg-Outdoor comprises \textbf{1,700 retained
source--receiver observations} from \textbf{370 source
groups} across \textbf{51 sequence windows}.
Each source group is associated with up to 12
candidate receiver poses.

\paragraph{Outdoor Evaluation References.}
AudLiSeg-Outdoor contains 286 valid test observations.
For the active-listening results reported in
Table~3 of the main paper, we require each reference
to support the complete five-position listening
budget, with feasible receiver poses and corresponding
four-channel acoustic observations.
Under this criterion, 266 references satisfy the
evaluation protocol (53 Near-Field, 104 Mid-Field,
and 109 Far-Field). The remaining 20 observations
retain valid source annotations but lack sufficient
multi-position acoustic coverage. All listening
budgets in Table~\ref{tab:outdoor_results} are evaluated on the same
reference subset to ensure consistent comparisons.

\paragraph{Outdoor Metric Aggregation.}
Outdoor metrics are computed independently for each
distance tier by averaging over individual evaluation
references. Exact source accuracy is the percentage
of references for which the predicted source instance
matches the ground truth.

\paragraph{Source Selection and Observation Retention.}
We use receiver poses from actual KITTI frames. Eligible vehicle
tracks must contain at least 30 annotated LiDAR points
in at least 12 frames within a sequence window.
For each eligible track, we select the window with the
largest number of valid frames and uniformly sample
12 receiver poses. Tracks with stronger source
visibility are prioritized while maintaining coverage
across sequence windows. This procedure initially
produces 4,800 outdoor candidate observations.
We subsequently retain source-receiver pairs with
distances not exceeding 13~m, yielding the
final 1,700 observations from 370 source groups.
This filtering is independent of rendered audio,
model predictions, segmentation quality, and
evaluation results.

\subsection{Dataset Statistics and Difficulty Tiers}
\label{supp_sec:dataset_statistics}

Table~\ref{supp_tab:dataset_stats} summarizes the composition of AudLiSeg. Together, the indoor and outdoor branches provide 6,500 observations spanning complementary environments, with consistent four-channel acoustic sensing and point-level source annotations.

\begin{table}[t]
\centering
\caption{\textbf{AudLiSeg dataset statistics.}
Dataset composition and training, validation, and test splits. Indoor and outdoor partitions are scene-disjoint and sequence-disjoint, respectively.}
\label{supp_tab:dataset_stats}

\small
\setlength{\tabcolsep}{7pt}
\renewcommand{\arraystretch}{1.2}

\resizebox{\columnwidth}{!}{%
\begin{tabular}{@{}lccrrrr@{}}
\toprule
& \multicolumn{2}{c}{\textbf{Dataset Scale}}
& \multicolumn{4}{c}{\textbf{Observations}} \\
\cmidrule(lr){2-3}
\cmidrule(l){4-7}
\textbf{Subset}
& \textbf{Scenes / Windows}
& \textbf{Sources / Groups}
& \textbf{Train}
& \textbf{Val.}
& \textbf{Test}
& \textbf{Total} \\
\midrule

Indoor
& 52 & 400
& 3,264 & 768 & 768
& \textbf{4,800} \\

Outdoor
& 51 & 370
& 1,088 & 326 & 286
& \textbf{1,700} \\

\midrule
\rowcolor{black!4}
\textbf{Combined}
& -- & --
& \textbf{4,352}
& \textbf{1,094}
& \textbf{1,054}
& \textbf{6,500} \\
\bottomrule
\end{tabular}%
}
\vspace{-10pt}
\end{table}

\begin{figure}[h!]
    \centering
    \includegraphics[width=1.0\columnwidth]
    {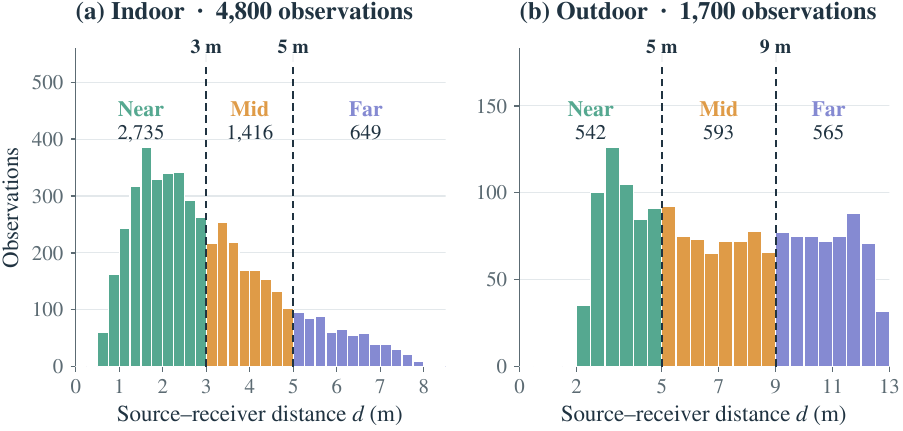}
    \caption{\textbf{Source--receiver distance distributions in AudLiSeg.}
    Histograms of (a) indoor and (b) outdoor
    observations, colored by distance tier.
    Dashed lines indicate the tier boundaries,
    and annotations report the observation counts.}
    \label{suppl_fig:distance_tiers}
\end{figure}

\vspace{5pt}
\noindent\textbf{Distance Distribution.}
Fig.~\ref{suppl_fig:distance_tiers} illustrates
the source--receiver distance distributions for
both benchmark branches. The indoor observations
are concentrated at shorter distances, whereas
the outdoor branch covers a wider spatial range.
The Near-, Mid-, and Far-Field partitions enable
evaluation across progressively larger
source--receiver distances.

\begin{figure}[h]
    \centering
    \includegraphics[width=1.0\columnwidth]
    {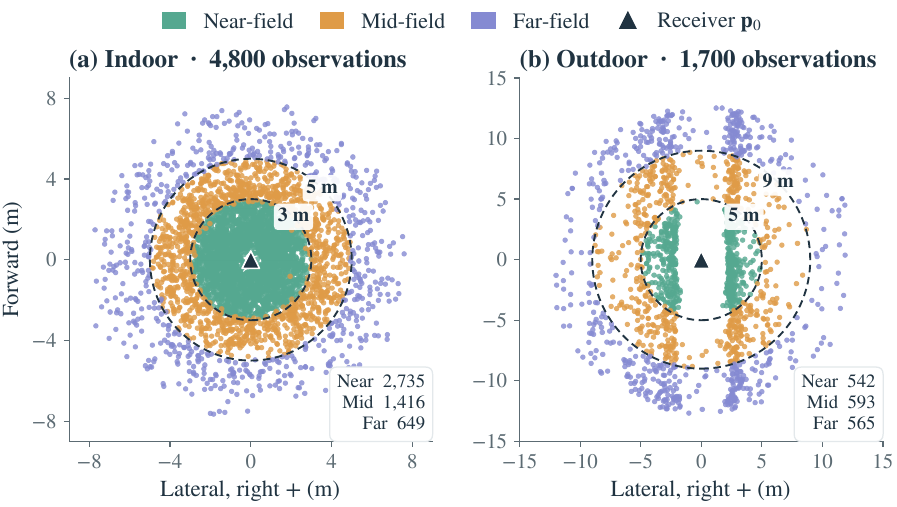}
    \caption{\textbf{Spatial distribution of sound sources in AudLiSeg.}
    Receiver-centered source locations for
    (a) indoor and (b) outdoor observations.
    Colors indicate the distance tiers, dashed
    boundaries show the corresponding thresholds,
    and the triangle denotes the reference
    receiver pose $\mathbf{p}_0$.}
    \label{suppl_fig:source_layout}
\end{figure}

\paragraph{Spatial Source Distribution.}
Fig.~\ref{suppl_fig:source_layout} visualizes
the spatial distribution of sound sources
relative to the reference receiver pose.
Indoor sources exhibit a broadly distributed
spatial arrangement, whereas outdoor sources
show a more elongated distribution consistent
with the geometry of driving environments.
Both branches cover multiple source directions
and distance tiers.

\begin{figure}[h]
    \centering
    \includegraphics[width=1.0\columnwidth]
    {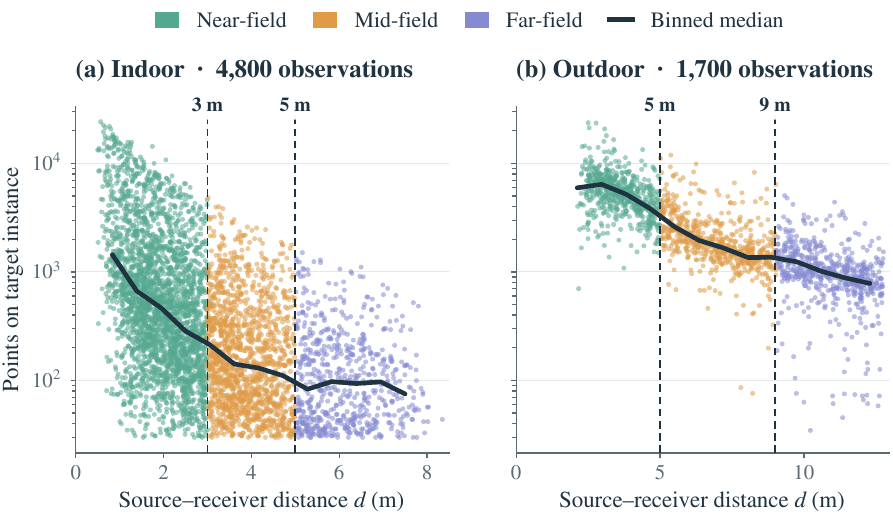}
    \caption{\textbf{Target-instance LiDAR point counts across distance.}
    Number of points belonging to each active
    source instance versus source--receiver
    distance for (a) indoor and (b) outdoor
    observations. Colors indicate distance
    tiers, and black curves show the binned
    medians. The vertical axis uses a
    logarithmic scale.}
    \label{suppl_fig:target_points}
\end{figure}

\paragraph{Target Point Support.}
Fig.~\ref{suppl_fig:target_points} examines
the number of LiDAR points belonging to the
active source instance as a function of
source--receiver distance. Both branches
exhibit a general reduction in target-point
counts at larger distances, although substantial
variation exists across observations.
This trend highlights an additional challenge
of distant sound-source segmentation, where
the target may be represented by fewer LiDAR
returns.

\paragraph{Distance-Based Difficulty.}
We divide observations into \textbf{Near-Field}, \textbf{Mid-Field}, and \textbf{Far-Field} tiers according to the source--receiver distance $d$. The thresholds are adapted to the different spatial scales of the indoor and outdoor environments. These tiers characterize the acoustic observation distance rather than different sensing modalities or independently collected datasets. 
Fig.~\ref{suppl_fig:distance_tiers} reports their distributions and distance ranges.

\begin{figure*}[t]
    \centering
    \includegraphics[width=0.90\textwidth]{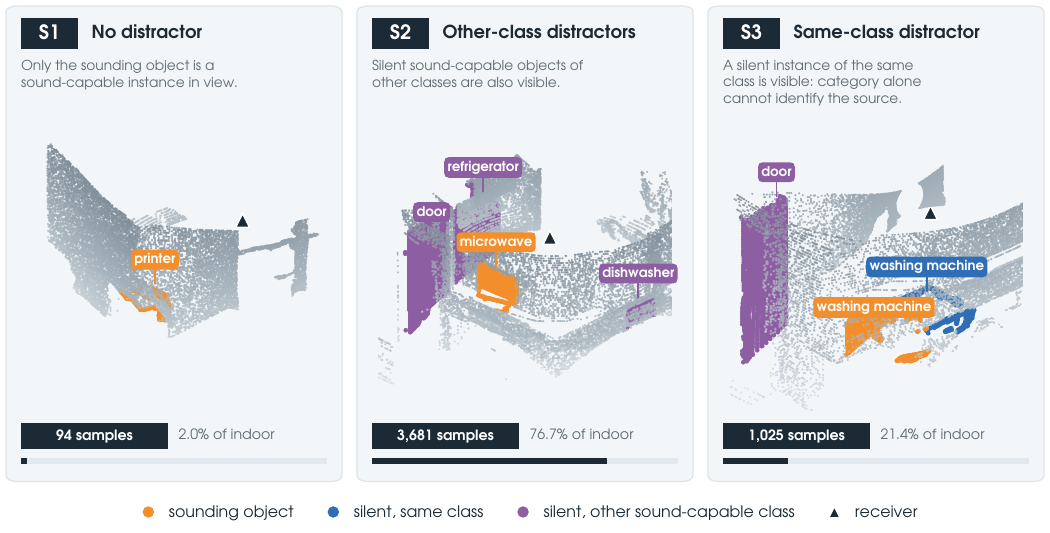}
    \caption{\textbf{Instance ambiguity settings in AudLiSeg-Indoor.}
    Representative examples of the three distractor
    settings: (S1) no sound-capable distractors,
    (S2) silent distractors from different semantic
    categories, and (S3) at least one silent
    same-category distractor. Each observation
    contains exactly one active source.
    The corresponding sample counts and proportions
    are shown for each setting.}
    \label{suppl_fig:distractor_settings}
\end{figure*}

\paragraph{Instance Ambiguity.}
Beyond distance-based difficulty, AudLiSeg-Indoor
characterizes instance-level ambiguity according to
visible sound-capable distractors, as illustrated in
Fig.~\ref{suppl_fig:distractor_settings}.
We consider three settings: \textbf{S1}, with no
distractors (94 observations); \textbf{S2}, containing
only different-category distractors (3,681 observations);
and \textbf{S3}, containing at least one same-category
distractor (1,025 observations).
While S1 and S2 primarily evaluate source grounding
under increasing semantic competition, S3 requires
distinguishing the active source from visually similar
instances of the same category.
Importantly, each observation contains exactly one
active sound source, making these settings a measure
of instance-level acoustic ambiguity rather than
simultaneous source separation.

\section{ActListen Implementation Details}
\label{supp_sec:actlisten}

This section provides additional implementation details
for the acoustic representations, geometry-conditioned
world model, predictive listening, and training procedure.
The overall ActListen formulation and inference process
are described in Sec.~4 of the main paper.

\subsection{Spatial Acoustic Evidence}
\label{supp_sec:acoustic_evidence}

\paragraph{Audio Preprocessing and Band Activity.}
Four-channel audio is processed at $44.1$~kHz using
a short-time Fourier transform (STFT) with a
4096-sample Hann window and a 2048-sample hop.
Frequencies between $150$ and $3000$~Hz are
partitioned into $B=48$ equal-width bands.
For the STFT representation
$\mathbf X(f,\tau)\in\mathbb C^4$, the spatial
covariance and its trace-normalized form are
\begin{equation}
\mathbf C_f=
\frac{1}{T_f}\sum_{\tau}
\mathbf X(f,\tau)\mathbf X(f,\tau)^{\mathsf H},
\qquad
\bar{\mathbf C}_f=
\frac{\mathbf C_f}
{\max(\operatorname{tr}\mathbf C_f,10^{-20})}.
\label{supp_eq:covariance}
\end{equation}
Normalized covariances are averaged within each
frequency band to obtain
$\mathbf Q^{\mathrm{obs}}_{tb}\in\mathbb C^{4\times4}$.
Trace normalization suppresses absolute source-power
variations while preserving inter-microphone
spatial relationships.

To identify informative frequency regions,
we compute band energy $e_{tb}$ from the traces
of the unnormalized covariance matrices.
The active-band indicator is defined as
\begin{equation}
m_{tb}=
\mathbb I\left[
e_{tb}>
\max\left(
10^{-4}\max_j e_{tj},10^{-20}
\right)
\right].
\label{supp_eq:active_band}
\end{equation}
Active bands contribute to acoustic matching,
while inactive bands are excluded from measured
evidence. When at least one band is active, their
normalized reliability weights are
$r_{tb}=m_{tb}/\sum_jm_{tj}$.

\paragraph{Geometry-Conditioned Acoustic Prediction.}
The acoustic world model combines an analytical
direct-path component with a learned covariance
prediction. For a source position $\mathbf s$
and microphone position $\boldsymbol\delta_m$,
let $d_m=\|\mathbf s-\boldsymbol\delta_m\|_2$.
The direct-path transfer function is
\begin{equation}
h_m(f)=\frac{1}{d_m}
\exp\left[
-\mathrm j\frac{2\pi f}{c}(d_m-d_1)
\right],
\label{supp_eq:direct_transfer}
\end{equation}
where $c=343$~m/s denotes the sound speed.
The normalized outer product
$\mathbf h(f)\mathbf h(f)^{\mathsf H}$
is pooled into the corresponding frequency bands.

The learned component predicts a complex
$4\times2$ factor $\mathbf L_b$ and constructs
$\mathbf Q_b^{\mathrm{learn}}
=\mathbf L_b\mathbf L_b^{\mathsf H}
+0.01\mathbf I$, followed by trace normalization.
The final covariance is
\begin{equation}
\mathbf Q_b=
(1-\gamma_b)\mathbf Q_b^{\mathrm{dir}}
+\gamma_b\mathbf Q_b^{\mathrm{learn}},
\qquad
\gamma_b=\sigma(v_b-1).
\label{supp_eq:covariance_mixture}
\end{equation}
This formulation preserves positive semidefiniteness
while allowing geometry-dependent deviations from
analytical propagation. The decoder additionally
predicts a bandwise log-error scale
$\ell_b=-2+2\tanh(u_b/2)$.

\paragraph{Geometric Representation and Architecture.}
The acoustic model uses measured LiDAR geometry
and frozen Utonia point features.
The viewing sphere is divided into
$48$ azimuthal and $4$ elevation sectors,
retaining the nearest and farthest measured
points in each occupied sector. This produces
at most 384 geometric tokens.

Global scene features are obtained through
mean and max pooling. For each candidate
emission hypothesis, separate attention
operations aggregate geometry local to the
candidate and along the source--receiver
segment. The resulting representations
initialize a 128-dimensional acoustic state.

The covariance decoder predicts the 48-band
acoustic representation and corresponding
error scales. A recurrent unit assimilates
acoustic residuals, while a separate recurrent
unit propagates the latent state under
receiver transformations. Both operate
with 128-dimensional hidden states.

\subsection{Closed-Loop Predictive Listening}
\label{supp_sec:closed_loop_listening}

\paragraph{Acoustic Feature Projection.}
For predictive acoustic imagination, each
$4\times4$ Hermitian covariance matrix is
represented by 16 real coordinates.
The off-diagonal components are scaled by
$\sqrt{2}$ to preserve Frobenius distance
under Euclidean vectorization.

Concatenating all 48 frequency bands produces
a 768-dimensional acoustic representation.
Principal component analysis (PCA) reduces
this representation to 128 dimensions.
PCA is fitted using training-only acoustic
observations, without access to validation
or test data.

Inactive measured frequency bands are masked
before evidence evaluation. During recurrent
memory assimilation, they are replaced by
the corresponding model predictions to avoid
introducing unsupported acoustic residuals.

\paragraph{Next-Best-Ear Planning.}
Next-Best-Ear (NBE) evaluates candidate
receiver actions using hypothetical acoustic
observations and the expected change in
posterior-weighted mask utility, as formulated
in Sec.~4.3 of the main paper. The candidate-overlap matrix distinguishes ambiguity between similar mask extents from ambiguity between different objects. Confusing
nearly identical masks incurs less overlap loss than selecting a
disjoint object. For mutually disjoint, nonempty candidate masks,
the utility reduces to the expected change in the maximum candidate
posterior probability; if all candidates have identical nonempty
support, it is zero. Both cases use the MAP mask decision applied
at inference.

For each candidate action, we draw
$S=16$ imagined acoustic observations.
The cross-entropy method (CEM) searches
over feasible ground-plane positions and
receiver yaw, evaluating 64 candidate poses
per listening decision. Common random draws
are reused across candidate actions to
reduce variance in utility comparisons.

Only the highest-utility action is executed.
Imagined observations are used exclusively
for planning and do not modify the retained
belief or acoustic memory. The listening
budget counts only acquired recordings,
not hypothetical observations.

\subsection{Evidence and Memory Feedback}
\label{supp_sec:evidence_memory_feedback}

\paragraph{Belief and Memory Updates.}
Using the notation of Eq.~(7) in the main paper, the acquired
observation updates the candidate and conditional site beliefs as
\begin{equation}
\begin{aligned}
p_{t+1}(k)&\propto p_t(k)\exp(d_k),\\
\rho_{t+1}(h\mid k)
&=\rho_t(h\mid k)\exp(\lambda_{kh}-c_k).
\end{aligned}
\label{supp_eq:belief_marginals_update}
\end{equation}
The common correction $d_k-c_k$ calibrates the candidate marginal
without changing the relative site evidence supplied by $\lambda_{kh}$.
With unit evidence scale and no neural score correction, $d_k=c_k$,
recovering the component-likelihood update. The previous posterior
supplies the prior, so initial candidate scores are not added again.

Acoustic memory is updated using the
shared channel-wise gate in Eq.~(8)
of the main paper. The gate is initialized
to recover the original recurrent
assimilation update and is subsequently
learned from acoustic sequences.
When an acquired recording contains no
usable acoustic evidence, the source
belief remains unchanged while the
action-predicted state is propagated.

\subsection{Training}
\label{supp_sec:training}

\paragraph{Staged Optimization.}
ActListen is trained in stages. After training
the perception frontend and base acoustic
model, these components remain frozen during
the optimization of the evidence adapter,
observation model, and memory gate.
All components are trained using the
corresponding training partitions, with
validation data used for model selection.
Ground-truth source annotations are not
available during inference.




\paragraph{Base Acoustic Model.}
Sec.~\ref{supp_sec:base_acoustic_model} specifies the base model's
trainable components, RIR and waveform supervision, and objective
over an initial observation and two receiver transitions. The model
selected by validation acoustic prediction error is frozen for the
subsequent stages below.

\paragraph{Evidence Adapter Training.}
The evidence adapter learns candidate-level
scores using soft segmentation-quality
targets proportional to
$\operatorname{IoU}
(\mathcal M_k,\mathcal M^*)^4$,
normalized across candidate masks.
A soft-target cross-entropy objective
encourages higher scores for candidates
with greater ground-truth overlap.

The adapter is optimized using AdamW.
Its calibration configuration and
likelihood scale are selected according
to validation segmentation performance.
The selected adapter is held fixed
during subsequent observation-model
training.

\paragraph{Observation Model Training.}
The observation model refines the
predictive acoustic distribution using
three complementary objectives:
\begin{equation}
\mathcal L_{\mathrm{obs}}=
\mathcal L_{\mathrm{NLL}}
+0.2D_{\mathrm{KL}}\left(
p_{\mathrm{real}}\|
p_{\mathrm{imag}}
\right)
+0.02\mathcal L_{\mathrm{corr}}.
\label{supp_eq:observation_world_loss}
\end{equation}
Here, $\mathcal L_{\mathrm{NLL}}$
measures the negative log likelihood
of the observed acoustic features
under the refined predictive mixture.
The KL term aligns candidate
distributions obtained from measured
and imagined acoustic representations,
using the frozen evidence adapter.
The correction regularizer penalizes
excessive predictive mean adjustments.

The observation model is optimized
using AdamW, with checkpoint selection
based on validation segmentation IoU
under the selected listening actions.
The base acoustic model and evidence
adapter remain frozen.

\paragraph{Memory Gate Training.}
The shared memory gate is trained
separately from sequences of observed
spatial acoustic features.
Training sequences are selected using
ground-truth source correspondence
to identify suitable candidate masks
and emission hypotheses. This
supervision is restricted to training
and validation.
Each sequence contains an initial
listening observation followed by
additional receiver observations.
The gate is optimized to reduce
reliability-weighted future acoustic
prediction errors, with regularization
on the gate parameters.
The base acoustic model, recurrent
assimilation mechanism, evidence
adapter, and observation model remain
frozen during gate optimization.
Model selection uses validation
prediction error, and no test
observations are used for training
or checkpoint selection.
Finally, NBE action selection is
performed through prediction-based
planning rather than end-to-end
differentiation through the final
segmentation objective.

\subsection{Base Acoustic Model}
\label{supp_sec:base_acoustic_model}

The base acoustic model learns geometry-conditioned
spatial covariance prediction and recurrent acoustic
state propagation. It is trained independently before
the evidence adapter, observation model, and memory
gate. The acoustic geometry encoder, covariance decoder,
observation-assimilation cell, and action-transition
cell are jointly optimized, while pretrained Utonia
features and the analytical direct-path component
remain fixed.

\paragraph{Training and Supervision.}
Training uses the AudLiSeg-Indoor training partition, with acoustic
observations acquired at different receiver positions for the same
stationary sound source. Each sequence begins
with a reference observation followed by two
receiver transitions. Ground-truth source positions
and simulated room impulse responses (RIRs) are
used exclusively for training supervision and are
not required during inference.

The model initializes an acoustic state from the
reference geometry and hypothesized source position,
assimilates the initial recording, and predicts
spatial covariances at subsequent receiver poses.
RIR-derived covariances provide source-signal-independent
targets, while waveform-derived covariances supervise
the predicted observation uncertainty.

For listening position $t\in\{0,1,2\}$, the
RIR prediction error is
\begin{equation}
E_t^{\mathrm{rir}}=
\frac{1}{B}\sum_{b=1}^{B}
\left\|
\mathbf Q_{tb}
-\mathbf Q_{tb}^{\mathrm{rir}}
\right\|_F^2,
\label{supp_eq:base_errors}
\end{equation}
where $B$ denotes the number of frequency bands.
The heteroscedastic observation loss is
\begin{equation}
H_t^{\mathrm{obs}}=
\frac{1}{B}\sum_{b=1}^{B}
\left[
\left\|
\mathbf Q_{tb}
-\mathbf Q_{tb}^{\mathrm{obs}}
\right\|_F^2
\exp(-\ell_{tb})
+\ell_{tb}
\right].
\label{supp_eq:base_heteroscedastic}
\end{equation}
The complete training objective is
\begin{equation}
\begin{aligned}
\mathcal L_{\mathrm{base}}
={}&E_0^{\mathrm{rir}}
+0.02H_0^{\mathrm{obs}}\\
&+\sum_{t=1}^{2}
\left(
0.5E_t^{\mathrm{rir}}
+0.01H_t^{\mathrm{obs}}
\right).
\end{aligned}
\label{supp_eq:base_training_loss}
\end{equation}
The RIR-based terms supervise geometry-conditioned
acoustic prediction, while the observation terms
calibrate uncertainty under variations in source
signals. The future-view losses additionally train
the recurrent state to preserve information relevant
to receiver motion.

\paragraph{Architecture and Optimization.}
The covariance decoder combines the latent acoustic
state with source-position features to predict
multiband spatial covariances and corresponding
log-error scales. An observation-assimilation module
updates the state using measured covariance residuals,
while a separate action-transition module propagates
the state under receiver transformations.
The candidate source selector is trained separately
from the base acoustic model and combines proposal
priors, semantic compatibility, and acoustic evidence
to initialize candidate beliefs. Its parameters are
not optimized by the base acoustic prediction loss.
The base model is optimized using AdamW, with
checkpoint selection based on validation acoustic
prediction error. Once trained, the base model
and source selector remain frozen during subsequent
ActListen training stages. Test observations are
excluded from training and checkpoint selection.

\section{Evaluation Protocol and Baselines}
\label{supp_sec:evaluation}

\subsection{Evaluation Metrics}
\label{supp_sec:metrics}

For AudLiSeg, predicted masks are evaluated on
the original reference LiDAR points, including
inverse mapping from the frontend's working
representation. Given a predicted foreground
mask $M$ and ground-truth mask $M^*$, pointwise
true positives (TP), false positives (FP), and
false negatives (FN) define
\begin{equation}
\begin{aligned}
\operatorname{IoU}
&=\frac{\mathrm{TP}}{\mathrm{TP+FP+FN}},\\
F_1
&=\frac{2\mathrm{TP}}
{2\mathrm{TP+FP+FN}}.
\end{aligned}
\label{supp_eq:metrics}
\end{equation}
Precision and recall are computed as
$\mathrm{TP}/(\mathrm{TP+FP})$ and
$\mathrm{TP}/(\mathrm{TP+FN})$, respectively.
Mean intersection-over-union (mIoU) and F1
are reported as percentages.

\paragraph{Source Identification Accuracy.}
Exact source accuracy measures whether the
selected candidate corresponds to the annotated
active physical object instance, rather than
merely matching its semantic category.
This distinction is particularly important
in scenes containing multiple sound-capable
objects of the same category.

For the comparison with SNAP, the threshold
metric evaluates the percentage of predictions
satisfying $\operatorname{IoU}>0.5$.
For qualitative analysis, segmentation outcomes
are grouped into \textit{Missed}
($\operatorname{IoU}=0$),
\textit{Partial}
($0<\operatorname{IoU}<0.5$), and
\textit{Good}
($\operatorname{IoU}\geq0.5$).

\paragraph{Metric Aggregation.}
For indoor active-listening evaluation,
results are reported separately for the
Near-Field, Mid-Field, and Far-Field
distance tiers. Following the main-paper
evaluation protocol, the Overall score is
the equal-weight macro-average across
these three tiers.

For any reported metric $z$,
\begin{equation}
z_{\mathrm{Overall}}
=
\frac{1}{3}
\left(
z_{\mathrm{Near}}
+z_{\mathrm{Mid}}
+z_{\mathrm{Far}}
\right).
\label{supp_eq:overall_aggregation}
\end{equation}
Thus, each distance tier contributes equally
to the Overall score, regardless of its
number of reference observations.
This aggregation differs from a
sample-count-weighted average over all test
references and from point-pooled evaluation.
In contrast, the single-view results (Table~\ref{tab:single_view_ablation} and Fig.~\ref{fig:teaser}\textcolor{cvprblue}{c} of the main paper)
are averaged over all 768 test references. Under the equal-tier
macro-average of Eq.~(20), the single-view mIoU is 38.75\%, so the gain from
one to five listening positions is $+6.71$ points on a like-for-like basis.

\paragraph{Receiver Travel and Statistical Evaluation.}
Cumulative receiver travel (\textbf{Dist.}) is measured
as the sum of shortest navigable-path lengths between
successive listening positions on the Habitat navigation
mesh. Each movement distance is computed using
Habitat's pathfinder geodesic distance, rather than
the Euclidean displacement between receiver poses.
Only executed movements contribute to the reported
cumulative travel; hypothetical candidate actions
evaluated during planning are excluded.

For uncertainty estimation, we use 10,000 matched scene-cluster bootstrap
resamples of the nine indoor test scenes. Each resample draws scenes with
replacement, retains all references from every drawn scene, and recomputes the
paired difference in mean mask IoU between the full model and each ablated
variant. Table~\ref{supp_tab:singleview_ci} reports the resulting 95\%
percentile intervals for the single-view ablations in Table~6 of the main
paper; all three intervals exclude zero.

\begin{table}[t]
\centering
\caption{\textbf{Scene-cluster bootstrap intervals for the single-view
ablations} (Table~\ref{tab:single_view_ablation} of the main paper). Differences are Full minus the ablated
variant, in mIoU percentage points.}
\label{supp_tab:singleview_ci}
\setlength{\tabcolsep}{4pt}
\resizebox{\columnwidth}{!}{%
\begin{tabular}{lcc}
\toprule
Comparison & $\Delta$mIoU & 95\% CI \\
\midrule
Full $-$ w/o Spatial acoustic evidence & 1.06 & [0.34, 1.84] \\
Full $-$ w/o Semantic evidence         & 0.95 & [0.28, 1.48] \\
Full $-$ Selector prior only           & 1.77 & [1.11, 2.32] \\
\bottomrule
\end{tabular}%
}
\vspace{-10pt}
\end{table}

\subsection{Active-Listening Evaluation Protocol}
\label{supp_sec:active_protocol}

Each active-listening episode begins with a reference
LiDAR scan and synchronized four-channel audio.
All methods use identical test references, initial
receiver poses, candidate masks, and navigability
constraints. A single sequential rollout is performed
per reference and method, with intermediate results
recorded for $T{+}1 \in \{2,3,4,5\}$ listening positions.
Near-, Mid-, and Far-Field assignments are fixed
according to the initial source--receiver distance.

\paragraph{Indoor Motion Constraints.}
Receiver movements are restricted to navigable
positions on the Habitat navigation mesh. Each
action is constrained to a maximum geodesic path
length of $6$~m from the current pose and a maximum
geodesic distance of $6$~m from the initial pose.
A minimum separation of $0.15$~m from previously
visited positions is maintained. The receiver height
is fixed at $1.2$~m. Subsequent positions contribute
only acoustic observations, without additional
LiDAR scans.

\subsection{Baseline Implementation Details}
\label{supp_sec:baseline_details}

We compare ActListen with four active-listening
alternatives under matched observation budgets
and a shared segmentation and acoustic evidence
pipeline. The comparison isolates the effect
of listening-position selection rather than
differences in LiDAR representations, candidate
generation, or acoustic evidence acquisition.

\paragraph{Prior-Forecast.}
Prior-Forecast retains the same candidate
belief, acoustic world model, Monte Carlo
planning objective, CEM search, and recurrent
evidence assimilation as ActListen.
However, it removes the learned observation
correction applied to future acoustic
predictions.
Consequently, hypothetical observations
are sampled from the calibrated base
acoustic forecast without the additional
learned correction.
This baseline isolates the contribution
of learned forecast refinement while
retaining world-model-based planning
and acoustic memory updates.

\paragraph{DirectValue.}
DirectValue predicts the utility of candidate
receiver actions directly from the current
scene and acoustic state, without sampling
imagined future observations.
Its input representation includes candidate
and emission-site acoustic predictions,
geometry and source-support features,
current source beliefs, latent acoustic
memory, residual information, and receiver
transforms.
Candidate-site representations are pooled
within each object hypothesis and aggregated
using posterior-weighted statistics across
the candidate bank. A learned value head
predicts the expected utility of each
proposed action.
The implementation is trained using a
combination of smooth-$L_1$ regression to
observed IoU improvement and a ranking
objective encouraging informative actions
to receive higher predicted scores.
Validation-based action selection determines
the retained checkpoint.
Unlike ActListen, DirectValue does not
generate hypothetical acoustic observations
or simulate their effect on the source
posterior. It therefore evaluates whether
direct action-value prediction can replace
explicit acoustic imagination under
comparable pre-acquisition information.

\paragraph{Analytical Planner.}
The analytical planner selects listening
positions using belief-weighted changes
in geometric bearing toward candidate
emission locations.
Rather than employing a learned action
utility or imagined acoustic evidence,
it evaluates receiver movements through
geometric changes relative to the current
source hypotheses.
This baseline tests whether a geometry-based
source-observation strategy is sufficient
for informative listening-position selection.

\paragraph{Random Pose.}
Random Pose samples a feasible receiver
position without evaluating predicted
acoustic informativeness.
It retains the same candidate masks,
acoustic evidence processing, belief
updates, and sensing budgets as the other
strategies. Thus, the difference is limited
to the selection of listening positions,
rather than the segmentation or
evidence-update mechanism.
For NBE-based methods, candidate actions
are evaluated through hypothetical
predictions before one action is committed.
Random Pose instead samples a feasible
position directly. In all cases, only
the selected position produces an
additional acquired audio recording.

\subsection{Comparison with EchoSegnet}
\label{supp_sec:echosegnet}

We evaluate the transferability of ActListen's
learned acoustic representations by integrating
them into the proposal-selection pipeline of
EchoSegnet~\cite{sokolov20243d}.
Unlike the primary AudLiSeg evaluation,
EchoSegnet operates on RGB--audio observations
and reconstructed 3D Gaussian scenes rather
than LiDAR-only geometry. To ensure a
controlled comparison, both methods retain
the same pretrained CAOD, SAM, ImageBind,
and 3D reconstruction components.

\paragraph{Adaptation Procedure.}
For each EchoSegnet object proposal, its
associated 3D Gaussian centers are used
to aggregate ActListen's acoustic predictions,
sequential source beliefs, and recurrent
world-model representations into a
26-dimensional acoustic feature vector.
These acoustic features are combined with
ImageBind similarity scores and proposal
statistics. A lightweight two-layer MLP
predicts a residual score for each object
proposal, refining the original audiovisual
proposal ranking.
All pretrained visual encoders and the
acoustic world model remain frozen.
Only the lightweight proposal-scoring
adaptation is optimized.

\paragraph{Evaluation Protocol.}
Evaluation is performed on the
3DAVS-S34-O7 dataset, comprising
34 scenes across four buildings.
We follow building-disjoint splits
and evaluate 90 held-out views.
Both methods use identical RGB--audio
inputs and 3D Gaussian reconstructions.
Following the benchmark protocol, we
report mask IoU, visible-target IoU,
and F-score.
This experiment examines whether ActListen's
acoustic representations improve source
grounding within an existing audiovisual
3D segmentation framework. It is a
separate cross-task adaptation experiment,
rather than an evaluation of the complete
LiDAR-only active-listening system.
Additional qualitative comparisons are
provided in Sec.~\ref{supp_sec:qualitative_echosegnet}.

\subsection{Comparison with Audio-Prompted SNAP}
\label{supp_sec:snap}

SNAP~\cite{gupta2026snap} supports
language-prompted point-cloud segmentation
but does not directly process acoustic
observations. To adapt SNAP for sound-source
grounding, we introduce an
audio-to-category-to-segmentation pipeline.

\paragraph{Adaptation Procedure.}
We initialize the SNAP-C decoder from
its pretrained checkpoint and fine-tune
it using AudLiSeg-Indoor point clouds
and corresponding semantic category masks.
At inference, a frozen CLAP model predicts
the sounding-object category from the
input audio by selecting among the
16 sound-capable categories.
The predicted category is converted
into a language prompt and encoded
using CLIP for SNAP-based segmentation.
We consider two variants:
\textit{SNAP + CLAP Category}, which
uses the audio-predicted semantic category,
and \textit{SNAP + GT Category}, which
substitutes the ground-truth category
as an oracle semantic prompt.
The oracle variant provides the correct
object category but does not reveal
the target instance identity or mask.
It therefore isolates the limitations
of category-guided segmentation beyond
semantic recognition errors.

\paragraph{Evaluation Protocol.}
We evaluate the SNAP adaptations on the
S1 and S2 ambiguity settings of
AudLiSeg-Indoor, where the target does
not compete with another visible
same-category sound-capable instance.
Both pipelines use identical LiDAR
point clouds. However, ActListen directly
uses the four-channel spatial audio,
whereas SNAP receives a language prompt
derived from the predicted or oracle
semantic category.
The comparison therefore evaluates
category-mediated point-cloud segmentation
against direct acoustic instance grounding,
rather than claiming identical acoustic
inputs across methods.
Additional qualitative comparisons are
presented in
Sec.~\ref{supp_sec:qualitative_snap}.

\section{Additional Results}
\label{supp_sec:additional_results}

We provide additional qualitative results of
\textbf{ActListen} across object categories,
source--receiver distances, and instance ambiguity
settings, together with comparisons against
SNAP and EchoSegnet. All LiDAR predictions use
the original reference point cloud without
additional geometric observations.

\subsection{Indoor Qualitative Results}
\label{supp_sec:qualitative_indoor}

\begin{figure*}[t]
    \centering
    \includegraphics[width=0.90\textwidth]{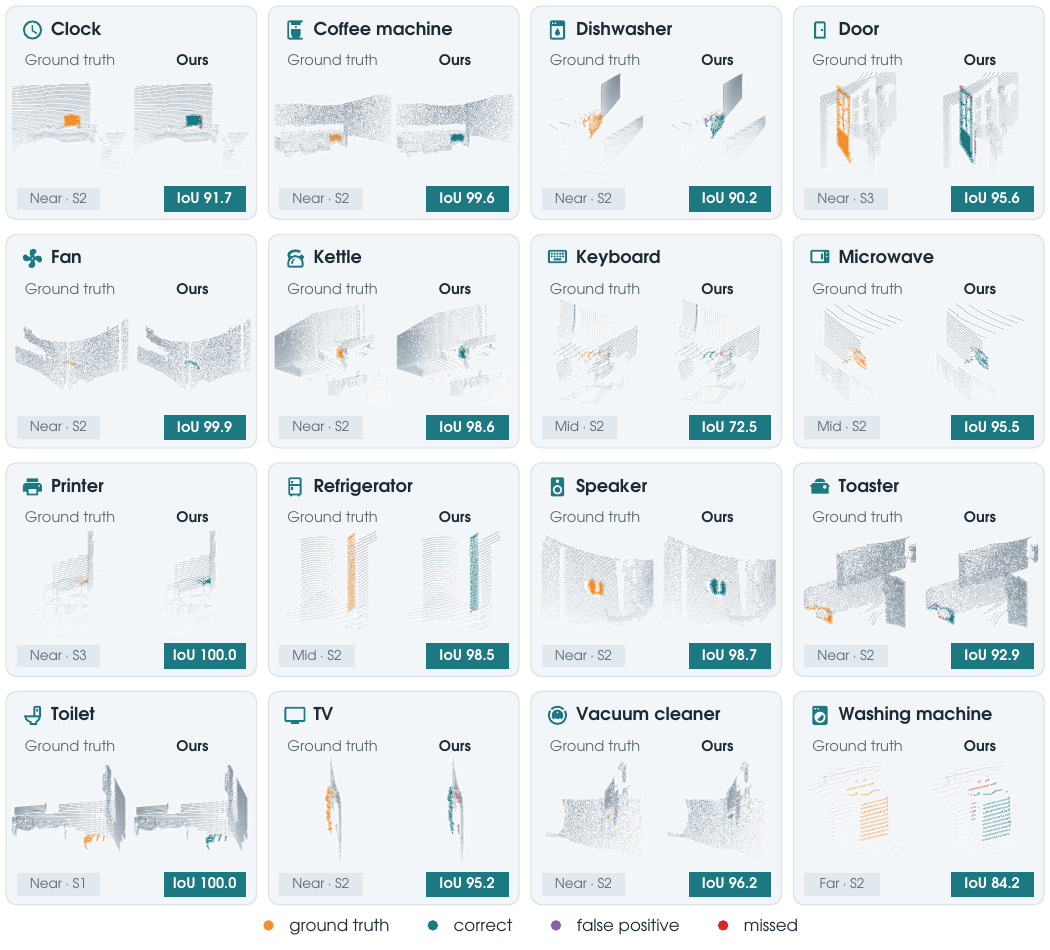}
    \caption{\textbf{Qualitative results across object
    categories on AudLiSeg-Indoor.}
    Representative ground-truth and ActListen
    segmentation results for all 16 sound-capable
    categories. Each example reports its distance tier,
    distractor setting, and mask IoU.
    Orange indicates the ground-truth target,
    teal denotes correctly predicted points,
    purple indicates false positives, and red
    indicates missed target points.}
    \label{suppl_fig:qual_indoor_class}
\end{figure*}

\begin{figure*}[t]
    \centering
    \includegraphics[width=0.90\textwidth]{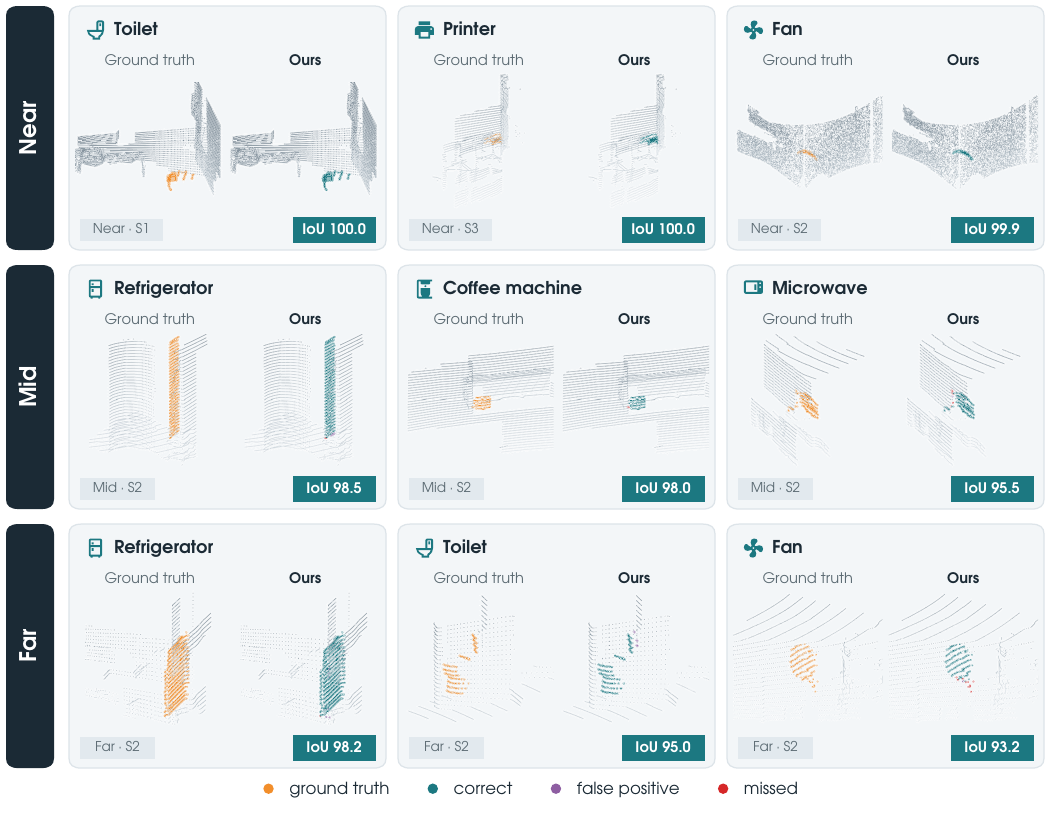}
    \caption{\textbf{Qualitative results across distance
    tiers on AudLiSeg-Indoor.}
    Representative ground-truth and predicted
    masks for Near-, Mid-, and Far-Field
    observations. Each panel reports the object
    category, distractor setting, and mask IoU.
    ActListen successfully segments selected
    sounding objects across different distances,
    including sparsely observed Far-Field targets.}
    \label{suppl_fig:qual_indoor_distance}
\end{figure*}

\begin{figure*}[t]
    \centering
    \includegraphics[width=0.90\textwidth]{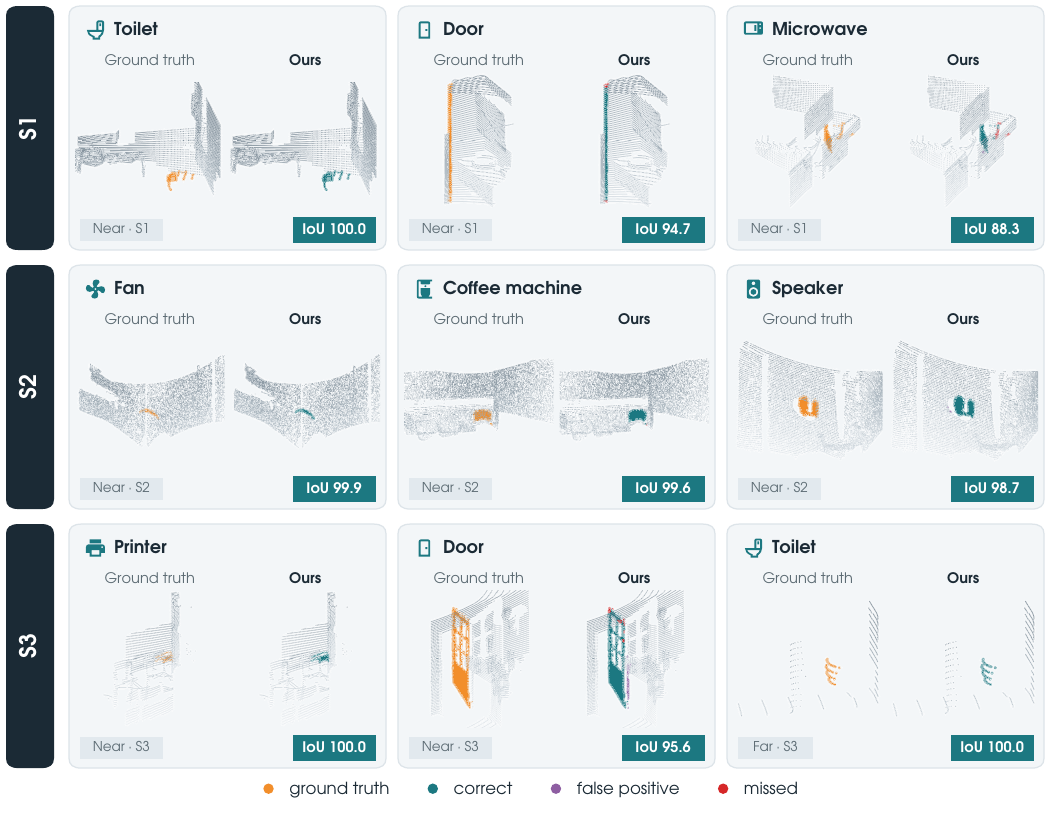}
    \caption{\textbf{Qualitative results across instance
    ambiguity settings on AudLiSeg-Indoor.}
    Representative examples from S1 (no distractors),
    S2 (different-category distractors), and
    S3 (same-category distractors).
    Ground-truth and predicted masks are shown
    alongside their IoU values. Selected S3
    examples demonstrate accurate identification
    of the active source despite the presence
    of semantically similar alternatives.}
    \label{suppl_fig:qual_indoor_setting}
\end{figure*}

Fig.~\ref{suppl_fig:qual_indoor_class} presents
representative results across all 16 sound-capable
categories in AudLiSeg-Indoor, illustrating
source grounding across diverse object geometries.
Fig.~\ref{suppl_fig:qual_indoor_distance}
shows successful predictions across Near-,
Mid-, and Far-Field settings, including
distant objects with sparse LiDAR observations.
Fig.~\ref{suppl_fig:qual_indoor_setting}
further examines the three instance ambiguity
settings (S1--S3), highlighting ActListen's
ability to distinguish sounding instances
even in the presence of same-category distractors.

\subsection{Progressive Active Listening}
\label{supp_sec:qualitative_progressive}

\begin{figure*}[t]
    \centering
    \includegraphics[width=0.95\textwidth]
    {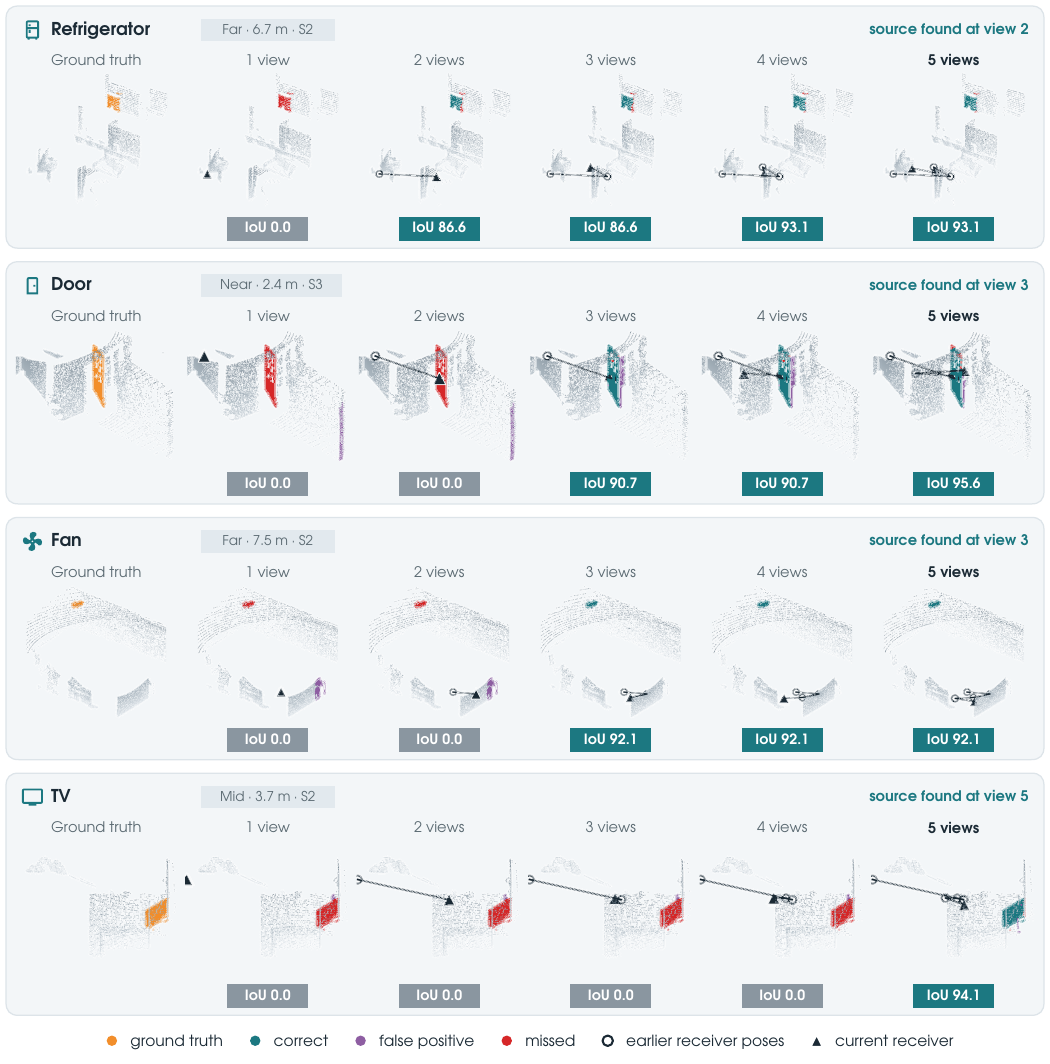}
    \caption{\textbf{Progressive sound-source segmentation
    through active listening on AudLiSeg-Indoor.}
    Four representative examples illustrate how additional
    listening positions recover initially missed sources.
    From top to bottom: refrigerator (identified at view 2),
    door (view 3), fan (view 3), and television (view 5).
    Each row shows the ground truth and predictions across
    five listening positions, with corresponding mask IoUs.
    Receiver markers indicate the current and previously
    visited positions. All predictions use the same reference
    LiDAR cloud without additional geometric observations.}
    \label{suppl_fig:qual_progressive}
\end{figure*}

Fig.~\ref{suppl_fig:qual_progressive}
illustrates four examples in which initially
missed sources are progressively recovered
through additional listening positions.
These improvements arise from sequential
acoustic evidence and updated source beliefs,
while the reference LiDAR geometry and
candidate masks remain fixed.

\subsection{Outdoor Qualitative Results}
\label{supp_sec:qualitative_outdoor}

\begin{figure*}[t]
    \centering
    \includegraphics[width=0.90\textwidth]{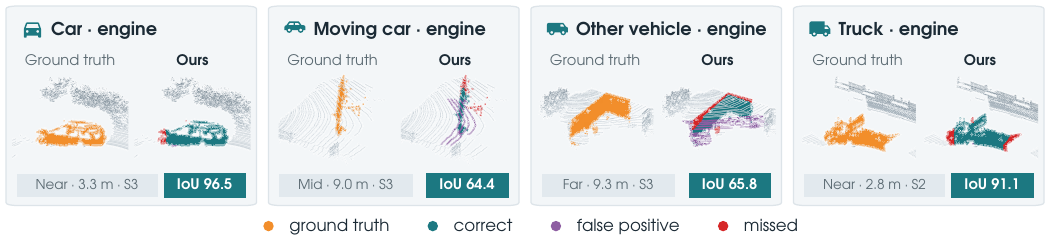}
    \caption{\textbf{Qualitative results across vehicle
    categories on AudLiSeg-Outdoor.}
    Representative ground-truth and predicted
    segmentations for car, moving-car, other-vehicle,
    and truck instances associated with engine sounds.
    Each example reports the source--receiver
    distance, distractor setting, and mask IoU.
    Correct predictions, false positives, and
    missed target points are distinguished
    by their corresponding colors.}
    \label{suppl_fig:qual_outdoor_class}
\end{figure*}

\begin{figure*}[t]
    \centering
    \includegraphics[width=0.90\textwidth]{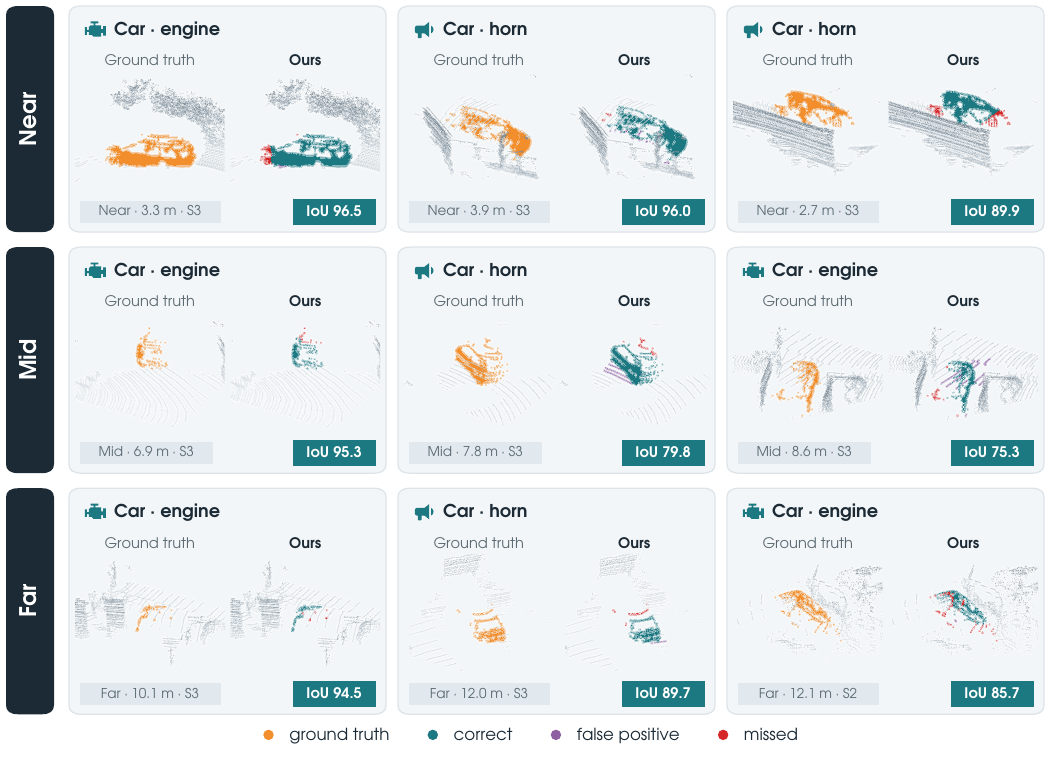}
    \caption{\textbf{Qualitative results across distance
    tiers on AudLiSeg-Outdoor.}
    Representative vehicle segmentation results
    for Near-, Mid-, and Far-Field observations
    involving engine and horn sounds.
    Each panel shows the ground-truth mask,
    ActListen prediction, source--receiver
    distance, distractor setting, and IoU.
    The selected examples demonstrate successful
    source grounding across varying target
    distances and LiDAR point densities.}
    \label{suppl_fig:qual_outdoor_distance}
\end{figure*}

Fig.~\ref{suppl_fig:qual_outdoor_class}
presents representative predictions across
the four vehicle categories in AudLiSeg-Outdoor.
Fig.~\ref{suppl_fig:qual_outdoor_distance}
further illustrates successful segmentation
across Near-, Mid-, and Far-Field settings.
While the selected examples demonstrate
accurate source grounding under varying
geometric conditions, distant and sparsely
observed vehicles remain challenging,
as reflected in the quantitative results.

\subsection{Qualitative Comparison with EchoSegnet}
\label{supp_sec:qualitative_echosegnet}

\begin{figure*}[t]
    \centering
    \includegraphics[width=0.95\textwidth]{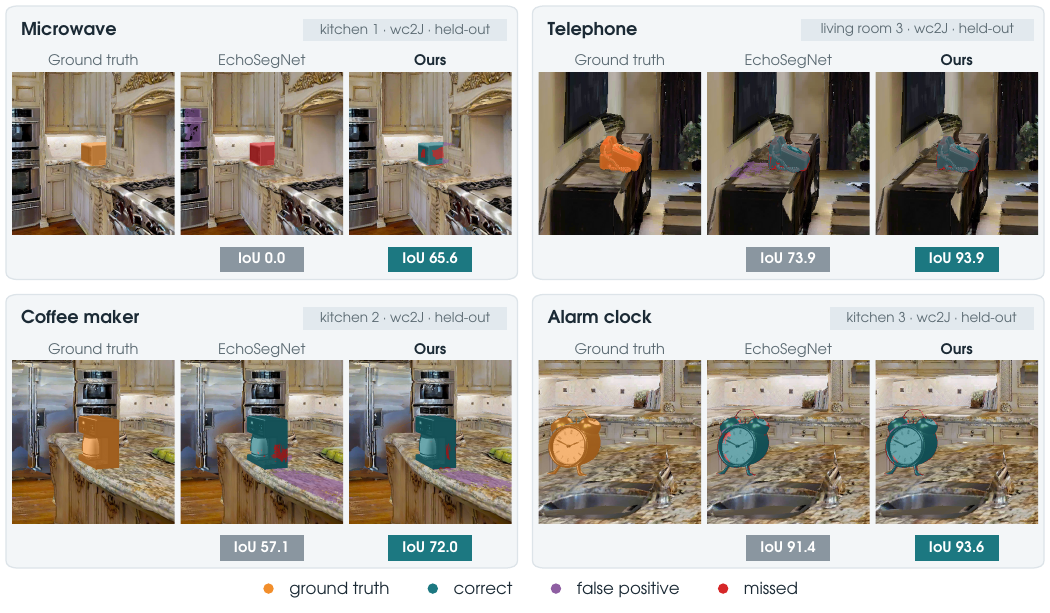}
    \caption{\textbf{Qualitative comparison with
    EchoSegnet on 3DAVS-S34-O7.}
    Ground-truth masks and predictions from
    EchoSegnet and the ActListen acoustic
    adaptation for four held-out examples.
    The comparisons illustrate improvements
    in target coverage and object association,
    with the corresponding mask IoU values
    reported for each example.}
    \label{suppl_fig:qual_echosegnet}
\end{figure*}

Fig.~\ref{suppl_fig:qual_echosegnet}
provides additional comparisons on
3DAVS-S34-O7. Integrating ActListen's
acoustic representations into the
EchoSegnet audiovisual pipeline improves
target coverage and reduces incorrect
object assignments in the selected examples.
This demonstrates the applicability of
the learned acoustic evidence beyond
LiDAR-only segmentation.

\subsection{Qualitative Comparison with SNAP}
\label{supp_sec:qualitative_snap}

\begin{figure*}[t]
    \centering
    \includegraphics[width=0.95\textwidth]{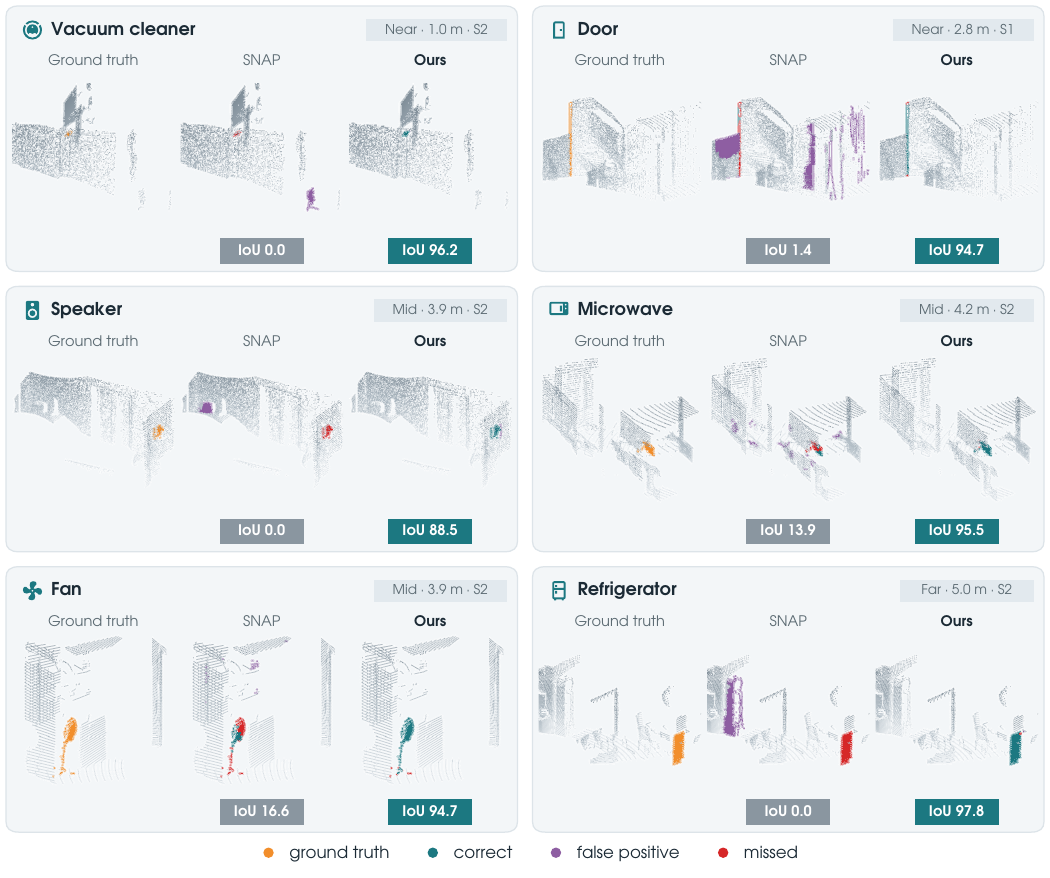}
    \caption{\textbf{Qualitative comparison with SNAP
    on AudLiSeg-Indoor.}
    Ground-truth masks and predictions from
    SNAP and ActListen across six object
    categories. The examples illustrate
    incorrect or incomplete grounding by
    SNAP and more accurate instance-level
    segmentation by ActListen.
    Mask IoU values are reported for
    each method and example.}
    \label{suppl_fig:qual_snap}
\end{figure*}

Fig.~\ref{suppl_fig:qual_snap}
compares ActListen with SNAP on
AudLiSeg-Indoor. In the selected examples,
ActListen more accurately identifies and
segments sounding instances, whereas SNAP
frequently produces incomplete or incorrect
object masks. These comparisons illustrate
the benefit of spatial acoustic grounding
over category-based prompting alone.

\section{Limitations and Discussion}
\label{supp_sec:limitations}

\subsection{Acoustic Reasoning in Challenging Conditions}
\label{supp_sec:acoustic_ambiguity}


Acoustic source association is challenging when several visible objects
belong to the same sound-capable category or when the active source is
distant. ActListen predicts the evidence expected at candidate receiver
poses and refines its source belief after acquiring the selected
recording. Additional recordings yield the largest outdoor gain in
the Far-Field tier, although individual observations need not improve
mask IoU. The recurrent state retains earlier acoustic evidence to
condition subsequent forecasts and listening decisions.

\subsection{Analysis of Segmentation Errors}
\label{supp_sec:mask_quality}

\begin{figure*}[t]
    \centering
    \includegraphics[width=0.90\textwidth]
    {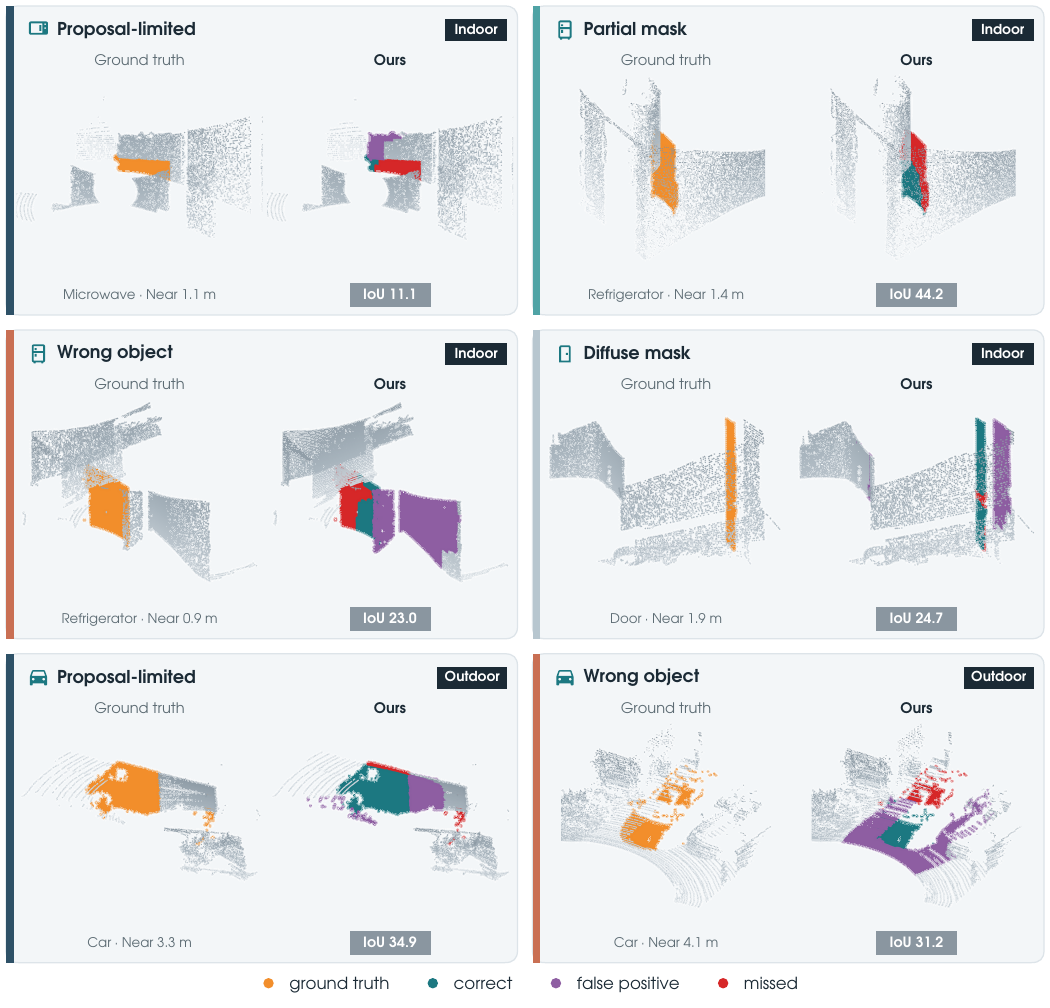}
    \caption{\textbf{Representative failure modes of
    ActListen on AudLiSeg-Indoor and AudLiSeg-Outdoor.}
    Examples illustrate proposal-limited segmentation,
    partial target coverage, incorrect object selection,
    and diffuse masks. Each panel compares the
    ground-truth target with the ActListen prediction
    and reports the corresponding mask IoU.
    Orange denotes ground truth, teal indicates
    correctly predicted points, purple represents
    false positives, and red denotes missed
    target points.}
    \label{suppl_fig:failure_modes}
\end{figure*}

Fig.~\ref{suppl_fig:failure_modes} presents representative
segmentation errors across indoor and outdoor environments,
illustrating the interaction between candidate-mask quality
and acoustic source association. The examples include
incomplete target coverage, diffuse predictions, and
incorrect object selection. For instance, the indoor
microwave and outdoor car examples exhibit limited
target coverage, whereas the refrigerator and door
examples show errors associated with competing object
regions or imprecise mask boundaries.

These observations are particularly relevant to ActListen's
fixed-geometry formulation. Since the framework selects
from candidate masks constructed using the initial LiDAR
scan, its achievable segmentation quality is bounded by
the best available candidate:
\begin{equation}
I_{\mathrm{bank}}^*
=
\max_k \operatorname{IoU}(\mathcal M_k,M^*),
\label{supp_eq:candidate_bound}
\end{equation}
where $M^*$ denotes the ground-truth mask.
This distinction separates candidate representation
quality from acoustic evidence-based selection.
Importantly, ActListen focuses on the latter:
it improves identification of the sounding instance
through additional acoustic observations without
requiring further LiDAR acquisition or geometric
reconstruction. The examples also motivate future
integration with adaptive candidate refinement,
which could complement the current active listening
mechanism by improving the available mask representations.

\subsection{Generalization and Future Directions}
\label{supp_sec:sim_to_real}

AudLiSeg-Indoor uses simulated geometry and acoustics, whereas
AudLiSeg-Outdoor combines measured LiDAR scans with synthesized audio.
Together, they provide a controlled setting for studying active acoustic
reasoning across scene geometries, source categories, and propagation
settings while keeping geometric observations fixed. The EchoSegnet
adaptation further evaluates the acoustic representation within an
existing RGB-based 3D audiovisual segmentation pipeline. Physical
microphone recordings, moving sources, and adaptive geometric
representations are complementary extensions of the modular framework;
their evaluation remains future work.

\clearpage

